%% file: main.tex
\documentclass[10pt,twocolumn,letterpaper]{article}

\usepackage[pagenumbers]{cvpr} 
\input{preamble}

\definecolor{cvprblue}{rgb}{0.21,0.49,0.74}
\usepackage[pagebackref,breaklinks,colorlinks,citecolor=cvprblue]{hyperref}

\usepackage[utf8]{inputenc} 
\usepackage[T1]{fontenc}    
\usepackage{hyperref}       
\usepackage{url}            
\usepackage{booktabs}       
\usepackage{amsfonts}       
\usepackage{nicefrac}       
\usepackage{microtype}      
\usepackage{enumitem}
\usepackage{graphicx}
\usepackage{multirow}
\usepackage{mathtools}
\usepackage{amsmath,amsfonts,amssymb,amsthm,bbm}
\usepackage{mathabx}
\usepackage{algorithm}
\usepackage{algorithmic}
\usepackage{colortbl}
\usepackage{amssymb}
\definecolor{mygray}{gray}{0.9}
\usepackage{wrapfig} 
\usepackage{caption}
\usepackage{pifont}

\newcommand{\vllm}{{Video-LLMs}}
\newcommand{\method}{{VTune}}

\newcommand{\cmark}{\ding{51}}%
\newcommand{\xmark}{\ding{55}}%

\def\paperID{195} 
\def\confName{CVPR}
\def\confYear{2025}

\title{On the Consistency of Video Large Language Models in Temporal Comprehension}

\author{%
  \bf Minjoon Jung\textsuperscript{\rm 1,2}\thanks{Work done during an internship at NUS}~~~
    Junbin Xiao\textsuperscript{\rm 1}\thanks{Corresponding author}~~~
    Byoung-Tak Zhang\textsuperscript{\rm 2}~~~ 
    Angela Yao\textsuperscript{\rm 1}\vspace{2mm}\\
   \textsuperscript{1}Department of Computer Science, National University of Singapore \vspace{0.5mm}\\
   \textsuperscript{2}Interdisciplinary Program in Artificial Intelligence, Seoul National University\vspace{1mm}\\
   \tt\small{\{mjjung,\,btzhang\}@bi.snu.ac.kr},~~~~~\tt\small{\{junbin,\,ayao\}@comp.nus.edu.sg}
   \vspace{-3mm}
}

\begin{document}
\maketitle
\input{secs/0.Abstract}
\input{secs/1.Introduction}
\input{secs/2.Related_Work}
\input{secs/3.Evaluation}
\input{secs/4.Experiments}
\input{secs/5.Solution_and_Discussion}
\input{secs/6.Conclusion}

{
    \small
    \bibliographystyle{ieeenat_fullname}
    \bibliography{main}
}

\input{secs/appendix}

\end{document}

%% file: preamble.tex
\usepackage[dvipsnames]{xcolor}

\usepackage{soul}


%% file: secs/0.Abstract.tex
\begin{abstract}
Video large language models (Video-LLMs) can temporally ground language queries and retrieve video moments. Yet, such temporal comprehension capabilities are neither well-studied nor understood. So we conduct a study on prediction consistency -- a key indicator for robustness and trustworthiness of temporal grounding. After the model identifies an initial moment within the video content, we apply a series of probes to check if the model's responses align with this initial grounding as an indicator of reliable comprehension. Our results reveal that current Video-LLMs are sensitive to variations in video contents, language queries, and task settings, unveiling severe deficiencies in maintaining consistency. We further explore common prompting and instruction-tuning methods as potential solutions, but find that their improvements are often unstable. To that end, we propose event temporal verification tuning that explicitly accounts for consistency, and demonstrate significant improvements for both grounding and consistency. Our data and code are open-sourced at \url{https://github.com/minjoong507/Consistency-of-Video-LLM}.
\end{abstract}
\vspace{-3mm}

%% file: secs/1.Introduction.tex
\section{Introduction}
\begin{figure}[t]
        \centering
        \includegraphics[width=0.99\linewidth]{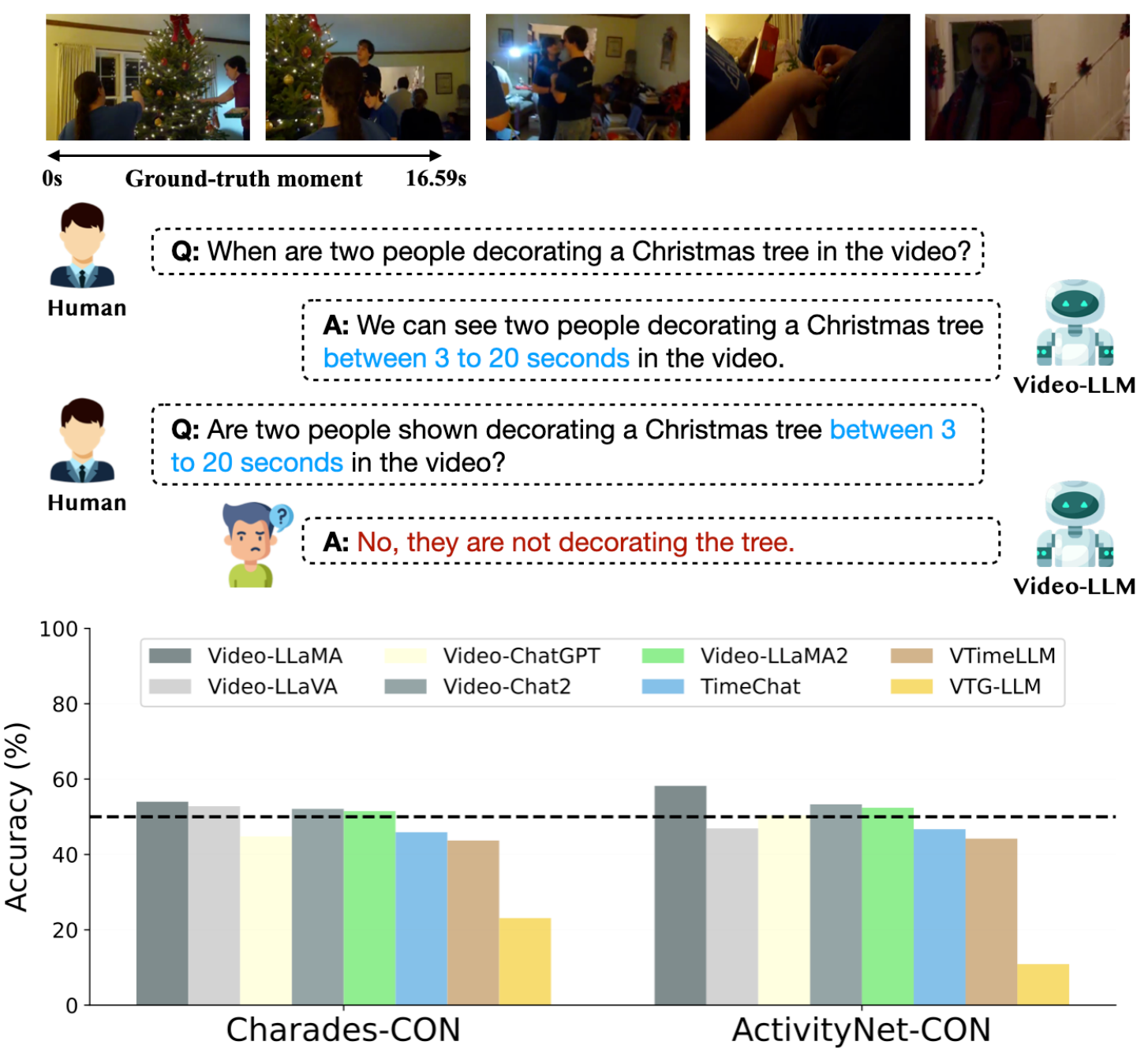}
    \caption{
    \textbf{Top}: An example of inconsistent behavior of \vllm, where the answer contradicts the initial temporal prediction in self-verification.
    \textbf{Bottom}: We reveal that most Video-LLMs struggle to reliably confirm their initial moment predictions, achieving a near chance-level consistency (50\%).
    }
    \label{fig: intro}
\end{figure}

Video large language models (\vllm)~\cite{videollama, lin2023videollava, li2023videochat, Maaz2023VideoChatGPT, cheng2024videollama2, luo2023valley} have shown great promise for video understanding. Yet they can follow only coarse-grained concepts and are challenged by identifying specific video moments for fine-grained comprehension. Recent works~\cite{timechat, vtimellm, qian2024momentor, guo2024vtgllm, wang2024hawkeye, huang2024lita, kahatapitiya2024language} try to improve temporal comprehension in \vllm~to handle time-related tasks, including video temporal grounding \cite{gao2017tall,krishna2017dense}, dense video captioning~\cite{krishna2017dense, zhou2018youcook2}, and grounded video question answering \cite{xiao2024can}. Despite their effectiveness, such capabilities of fine-grained temporal comprehension are not necessarily robust and trustworthy. For example, Figure \ref{fig: intro} shows that \vllm~are inconsistent when asked to check their own video moment predictions, raising concern about their capabilities in faithful temporal reasoning.

Consistency is a key indicator of robustness and trustworthiness, so we study how well current models maintain it when conversing about fine-grained video moments. Specifically, we focus on video temporal grounding, where the task involves identifying timestamps in a video that correspond to language queries. Intuitively, a model that links a language query to visual content in a specific temporal moment should remain robust to rephrasing queries and temporal shifts in the visual content.  We refer to this quality as \textbf{\emph{grounding consistency}}. Furthermore, if a model can identify a specific moment, its understanding should be bidirectional.  As such, the model should be able to confirm that the said event and its components did occur within the predicted moment. We refer to this quality as \textbf{\emph{verification consistency}}.

To evaluate such consistencies, we construct new evaluation sets, Charades-CON and ActivityNet-CON, based on two popular video temporal grounding datasets: Charades-STA \cite{gao2017tall} and ActivityNet-Captions \cite{krishna2017dense}. We curate test data by rephrasing the original queries and shifting the ground-truth moments in videos for probing \textbf{\emph{grounding consistency}}. To measure \textbf{\emph{verification consistency}}, we curate binary-choice questions to verify if the grounded moments truly reflect the queries and their key components. With the probes and corresponding data, we evaluate a range of prominent models, including two closed-source models: GPT-4o \cite{achiam2023gpt4} and Gemini \cite{reid2024gemini1.5} and eight open-source \vllm, including three `time-aware' models TimeChat~\cite{timechat}, VTimeLLM~\cite{vtimellm}, VTG-LLM~\cite{guo2024vtgllm} which have been developed specifically to grasp temporal moments.

Interestingly, we find that open-source models like Video-LLaVA~\cite{lin2023videollava} and Video-ChatGPT~\cite{Maaz2023VideoChatGPT} are often inconsistent in grounding video content and perform near chance-level when verifying their own answers. Despite their specialized designs, the time-aware models demonstrate limited improvements, particularly in verification consistency. Although the closed-source models may not be designed for temporal grounding, they demonstrate superior consistency, outperforming most open-source models. With the above observations, we further analyze two common solutions for \vllm~to improve consistency: prompting and instruction tuning with target data. We find that prompting cannot improve performance across different models. While instruction tuning consistently boosts grounding performance, it remains ineffective in improving consistency. Their improvements are often unstable, highlighting the challenges in enhancing the models' temporal comprehension.

To this end, we propose event temporal verification tuning (\method) - a method that extends instruction tuning by explicitly accounting for consistency. Rather than simply instructing the models to localize the video moment based on the query, we reformulate the task as a verification process. Specifically, we instruct the models to confirm aligned (rephrased) queries with positive moments. Furthermore, we prompt the model to identify and correct content changes by providing misaligned queries with positive moments and aligned queries with negative moments. By verifying such modified queries, we aim to enhance the model consistency while strengthening their original grounding capabilities. Our experiments show that \method~significantly improves both grounding and consistency across different models and datasets, demonstrating it to be a superior method for robust and trustworthy temporal comprehension. 

Our primary contributions are summarized as follows:
\begin{itemize}[leftmargin=*]
    \item We study the model's consistency in temporal comprehension by assessing whether its responses align with the initial grounding, using dedicated probes and datasets.

    \item We provide comprehensive consistency evaluation results, revealing significant deficiencies in \vllm~for robust and trustworthy temporal comprehension.

    \item We further share insights about the inadequacy of common prompting and instruction tuning solutions and propose \method~towards faithful temporal comprehension.
\end{itemize}

%% file: secs/2.Related_Work.tex
\section{Related Work}
\textbf{Video Temporal Grounding (VTG).} 
VTG requires fine-grained video-language understanding and sensitivity to temporal dynamics, similar to other tasks~\cite{lei2021qv, jung2022}. Existing VTG models can be categorized as proposal-based~\cite{wang2022mmn, zhang20202dtan}, proposal-free~\cite{zhang2020vslnet, mun2020local}, or transformer-based~\cite{moon2023qddetr, Jung_2025_WACV}. Proposal-based methods generate a set of candidate video moments with a sliding window and rank them by relevance to the query. Proposal-free methods directly regress timestamps within videos. Transformer-based methods are now popular solutions for temporal grounding; they adapt the detection capabilities of DETR~\cite{carion2020detr} to find video moments.
\begin{figure*}[t]
        \centering
        \includegraphics[width=1.0\linewidth]{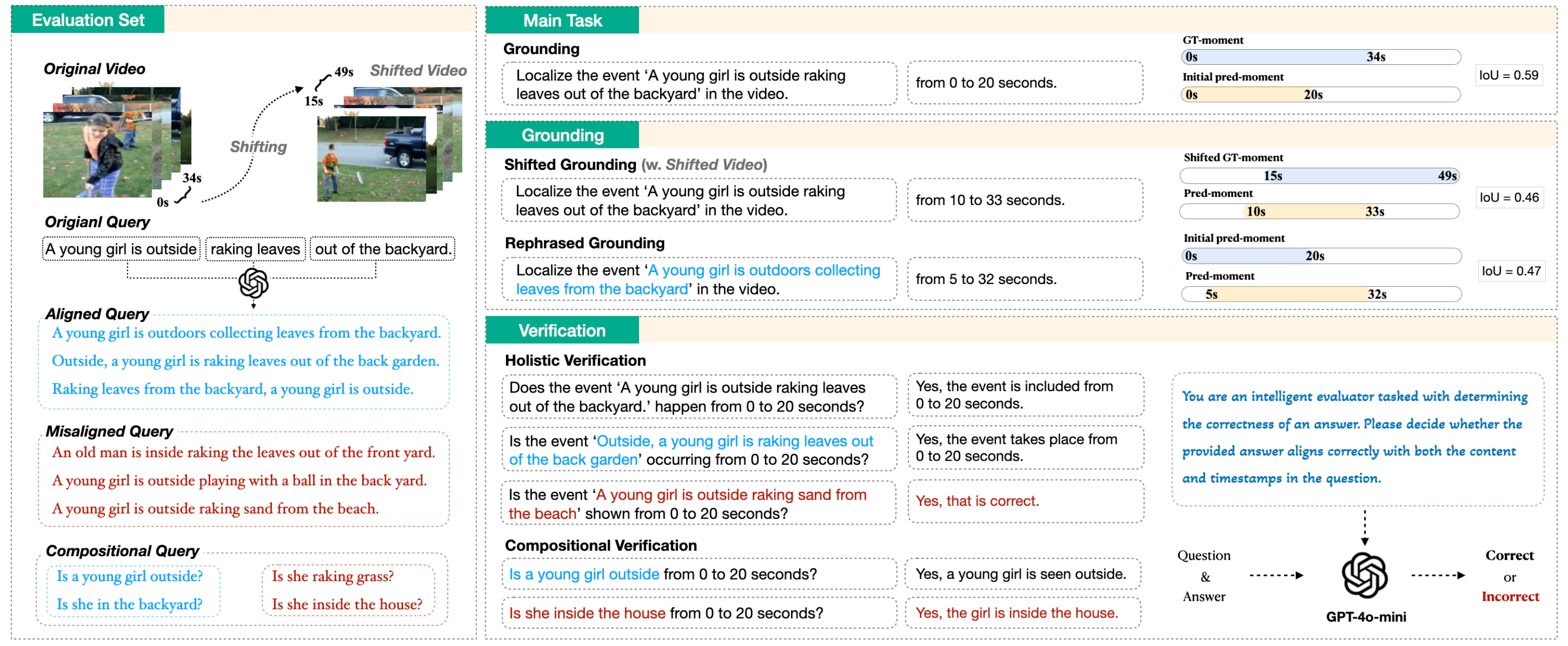}
    \caption{
    Illustration of our consistency evaluation process. For each query-moment pair in the video, we shift the ground-truth moment to a different moment and prompt GPT-4o-mini to generate aligned, misaligned, and compositional queries. We measure consistency as an IoU for grounding probes and design a GPT-based evaluation to assess the model's response for verification probes.
    }
    \label{fig: evaluation}
    \vspace{-3mm}
\end{figure*}

\noindent\textbf{\vllm~for VTG.}
\vllm~\cite{luo2023valley, li2024mvbench, videollama, lin2023videollava, Maaz2023VideoChatGPT, cheng2024videollama2} can chat about videos, but most are limited to coarse-grained understanding. A handful of recent \vllm~\cite{timechat, vtimellm, qian2024momentor, guo2024vtgllm, wang2024hawkeye} target grasping fine-grained temporal moments. Typically, these models are instruction-tuned to answer with timestamps to enhance their temporal comprehension capabilities. They have impressive progress in identifying moments or providing specific details in videos, but their strengths in temporal understanding are not well understood.  Specifically, is their performance truly grounded in video comprehension, or is it due to other spurious factors and correlations~\cite{otani2020uncovering, yuan2021closer, xiao2024can}? We thus analyze such capabilities by focusing on prediction consistency.

\noindent\textbf{Analysis of \vllm.} 
A line of works~\cite{li2024mvbench, liu2024tempcompass, li2023vitatecs, ning2023videobench, xiao2024videoqa} analyze and assess the temporal understanding of \vllm~from multiple aspects. However, they primarily focus on understanding model behaviors about temporal dynamics 
through question-answering rather than evaluating and improving model response consistency. Inconsistent behaviors in LLMs have been reported in previous studies for text~\cite{li2023benchmarking, wang2022selfcons, elazar2021measuring, zhou2022promptconcsistency} or image-text~\cite{yue2024sctune, zhang2024unveiling} understanding. Unlike prior efforts, we focus on the consistency of moment predictions in \vllm. Additionally, we explore and analyze potential solutions to improve consistency.

%% file: secs/3.Evaluation.tex
\section{Consistency Evaluation}
In this section, we describe the construction of evaluation sets (\S\ref{sec: dataset}) designed to support our probes (\S\ref{sec: probes}) and conclude with the evaluation configurations (\S\ref{sec:evaluation_configs}).

\subsection{Dataset Construction} 
\label{sec: dataset}
We construct Charades-CON and ActivityNet-CON by curating test sets from two popular video temporal grounding datasets, Charades-STA~\cite{gao2017tall} and ActivityNet-Captions~\cite{krishna2017dense}, respectively. These datasets mainly cover daily human activities with average video durations of 30 and 120 seconds, respectively. As temporal grounding datasets, they include labeled queries with corresponding timestamps for each video. We sample 500 videos from each test set and filter out the annotations where the query sentence or the moment is too short (\ie, less than 5 words or 5 seconds). We also exclude the moment annotations that exceed 70\% of the corresponding video length to avoid unreliable assessments, as models can simply predict the entire video length to recall these moments~\cite{yuan2021closer}. After filtering, there are 707 and 1,422 query-moment pairs for Charades-CON and ActivityNet-CON, respectively. We then use a strong LLM (\ie, GPT-4o-mini) to generate aligned, misaligned, and compositional queries to support our evaluation (please refer to Appendix Figure~\ref{fig: prompt to construct evaluation sets} for the prompt design). Queries are labeled as ``aligned'' or ``misaligned'' depending on whether they have the same or different meanings from the original query, respectively. Compositional queries partially overlap with the original by retaining some information, and we also generate them in both aligned and misaligned forms.

\noindent\textbf{Aligned \& Misaligned Queries.} Simply prompting GPT to rephrase the original queries often results in rephrased sentences too similar to the original. To introduce meaningful variations, we apply a set of carefully designed modification rules: (1) \textit{Word Replacement}: Substituting key nouns and verbs, (2) \textit{Active to Passive Conversion}: Changing active sentences to passive ones, and (3) \textit{Word Order Modification}: Rearranging the sentence structure. These transformations generate aligned queries intended to preserve the original query’s meaning while introducing differences in phrasing, aiming for a nuanced approach. For misaligned queries, we focus on slight alterations to key components. We prompt GPT to identify the main components in the original query and generate queries that are subtle yet irrelevant. Each query sentence is paired with three aligned and three misaligned queries to balance the distribution of ``Yes/No'' answers.

\noindent\textbf{Compositional Query.} We decompose query sentences down into three key components: (1) \textit{Subject Identification}: Identifying the main entities involved in the sentence, (2) \textit{Action}: Describing what the subjects are doing or what is happening to them, and (3) \textit{Relations}: Determining the relationship between the subjects and other elements. For example, given the query, ``A young girl is outside raking leaves out of the backyard,'' the model identifies ``a young girl'' as the subject, ``raking'' as the action, and associates ``outside'' and ``backyard'' as relational details. These components are converted into questions like ``Is a young girl outside?'' that ask for sub-information about video moments. Similarly, we generate misaligned ones by altering the components (\eg, ``Is she inside the house?''). On average, each query has 3.9 and 4.3 compositional queries in Charades-CON and ActivityNet-CON, respectively.

\noindent\textbf{Human Study.} Lastly, we randomly sample 700 question-moment pairs from the datasets, which have 2,100 aligned, misaligned, and compositional queries. We then ask humans to evaluate whether each sentence meets our expectations on a scale of 1 to 3—(1) Not matched, (2) Matched, and (3) Well-matched. If a misaligned query indeed unmatches the video content, it will be assigned a score of 3. Annotators independently evaluate the queries, and any ambiguities raised are cross-validated. The evaluators rate 92.2\% of the sentences with a score of 3 and 6.4\% with a score of 2, demonstrating the high quality of the curated data.

\begin{figure*}[t]
        \centering
        \includegraphics[width=1.0\linewidth]{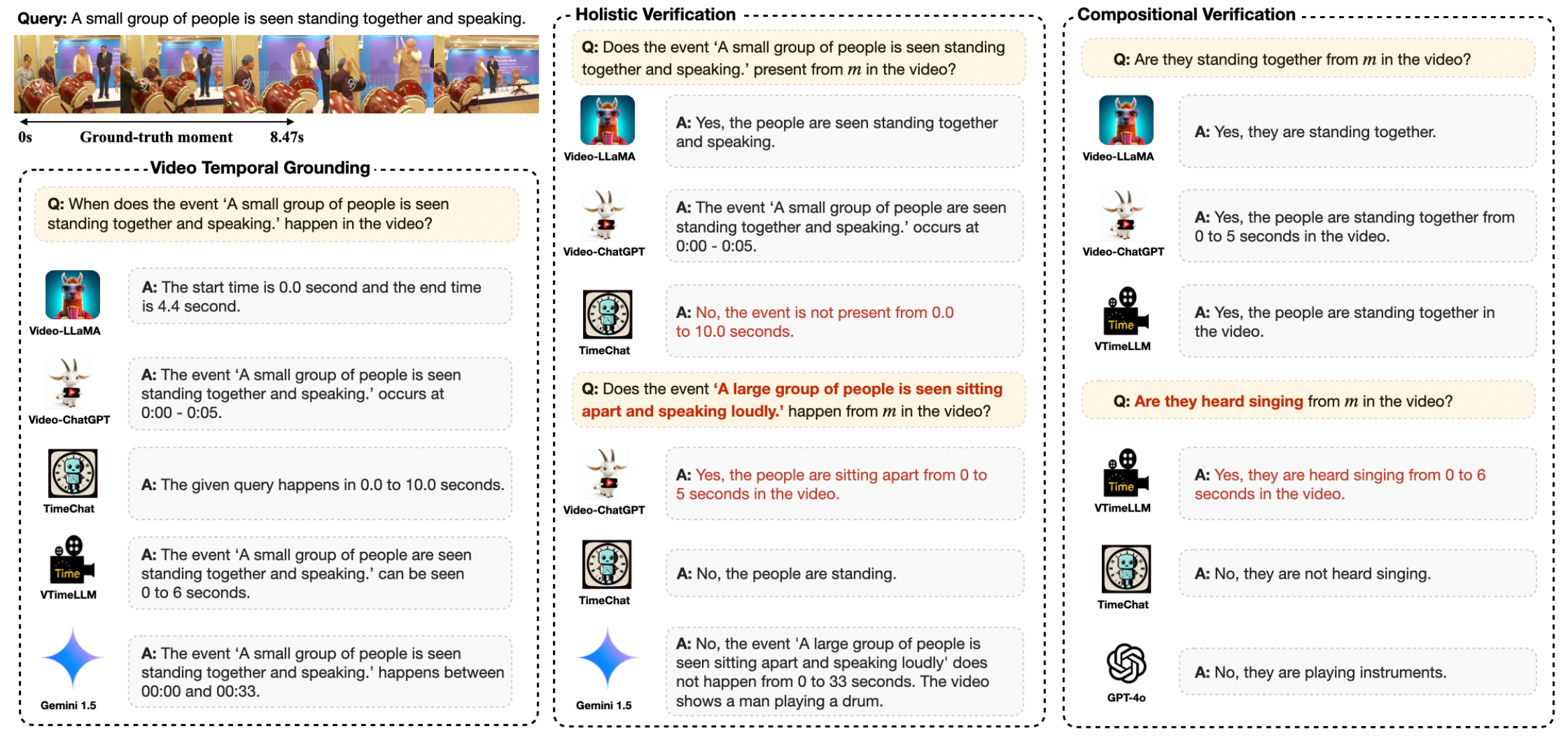}
    \caption{
    Examples of the model responses for verification probes. We first ask the model to predict the timestamp of the given sentence, then query it based on its own predictions. For holistic and compositional verifications, we replace the $m$ in the questions with each model's moment prediction. The \textcolor{red}{red} text indicates misaligned queries or highlights inconsistent model responses.
    }
    \label{fig: visualization}
    \vspace{-5mm}
\end{figure*}
\subsection{Consistency Probes} \label{sec: probes}
Consider a video $v$, a query sentence $q$, and its ground-truth moment as $m_{gt} \in \{t_s, t_e\}$ defined by start and end timestamps in $v$ respectively. The temporal grounding task (Ground) is defined as
\begin{equation}
m = \text{Temp}_G(v, q),
\end{equation}
where $m$ is the model's predicted moment. Verifying if the query $q$ corresponds to the $m$ in $v$ can be defined as 
\begin{equation}
    a = \text{Temp}_V(v, q, m) \in \{\text{Yes}, \text{No}\},
\end{equation}
where $a$ is the model's answer to verification probes. Temp$_G$ and Temp$_V$ denote the same model for different tasks. To probe prediction consistency, we introduce two grounding probes: Shifted (S-Ground) and Rephrased Grounding (R-Ground) and two verification probes: Holistic (H-Verify) and Compositional Verification (C-Verify). The probes are illustrated in Figure \ref{fig: evaluation} and we detail them below.

\noindent\textbf{Rephrased Grounding} evaluates if the model's moment predictions are consistent across queries $q$ and its aligned variant $\tilde{q}$. Specifically, we measure the IoU between $m=\text{Temp}_{G}(v, q)$ and $\tilde{m}=\text{Temp}_{G}(v, \tilde{q})$. We consider three aligned queries and report the average of their IoU values to measure the alignment between the moments. 

\noindent\textbf{Shifted Grounding} measures if the model consistently grounds the same visual content even if the content is shifted to different temporal positions. Specifically, while preserving the sequence of frames within the ground-truth moment $m_{gt}$, we randomly shift it to a new moment, {$m_{s}$}. In our implementation, the shifted moments overlap their original moments only 20\% on average, allowing us to assess whether the model can adjust their predictions accordingly.

\noindent\textbf{Holistic Verification} confirms whether the model can confirm the presence of the query $q$ within the moment prediction $m$, aligning with its initial prediction. This task assesses whether the model can respond affirmatively to confirm that all aspects of the query are represented within the moment, \ie, $\text{Temp}_{V}(v, q, m) = $ ``Yes.'' To prevent artificially high consistency scores from blanket affirmative answers, we use misaligned queries $\bar{q}$ to test if the model correctly denies their occurrence, yielding responses of $\text{Temp}_V(v, \bar{q}, m) = $ ``No.'' Additionally, we vary the prompt templates, including templates like ``Is the event $q$ missing in $m$?'' to shift the correct answer ``Yes.'' to ``No.'' Templates designs for verification prompts are listed in Appendix Table~\ref{tbl: templates for holistic verification}.
\input{tables/consistency}

\noindent\textbf{Compositional Verification} verifies whether the model's moment predictions are based on compositional reasoning rather than shallow pattern recognition. Specifically, given $q$ and its predicted moment $m$, we ask if the model correctly confirms the compositional visual facts of $q$ within $m$. For example, if the model answers ``0 to 20 seconds.'' for the query ``A young girl is outside raking leaves out of the backyard'', we verify whether it accurately identifies essential components, such as the presence of a young girl, the outdoor setting, and the action of raking leaves, within 0 to 20 seconds in the video. To further assess the model's understanding, we also ask questions like ``Is she raking grass?'' or ``Is she inside the house?'' with irrelevant components.

\subsection{Evaluation Configuration}
\label{sec:evaluation_configs}
We focus on correctly grounded queries, as evaluating consistency on incorrect predictions is less meaningful. Otherwise, any consistency check would merely test blanket response patterns rather than its true comprehension of the content. Thus, we filter out inaccurate moment predictions with an IoU below 0.5 relative to their ground-truth moments, applying ``Recall@1, IoU=0.5'' metric. We use the same metric for grounding probes. Results with different IoU values (e.g., 0.7) are shown in Appendix Table~\ref{tbl: grounding and consistency across different IoUs}. For verification probes, we assess QA accuracy as a consistency score based on correct moment predictions. However, models sometimes provide indirect answers instead of simple ``Yes/No'' responses (e.g., responding ``The event $q$ does not happen in $m$.'' to ``Does the event $q$ happen in $m$?''). In such cases, we prompt GPT-4o-mini for evaluation; see Appendix Figure~\ref{appendix_fig: prompt for evaluation}. For clarity, we report both the \textbf{absolute consistency score} (consistent predictions across the whole test set) and the \textbf{relative consistency score} (normalized within accurately grounded predictions).

%% file: tables/consistency.tex
\begin{table*}[t]
    \resizebox{\linewidth}{!}{
      \begin{tabular}{lcccccccccccc}
      \toprule
      \multirow{2}{*}{\bf Method} & \multirow{2}{*}{\bf LLM} & \multirow{2}{*}{\bf \# Frames} & \multicolumn{5}{c}{\bf Charades-CON} & \multicolumn{5}{c}{\bf ActivityNet-CON} \\ 
       \cmidrule(lr){4-8} \cmidrule(lr){9-13} 
        & & & Ground & R-Ground & S-Ground & H-Verify & C-Verify
        & Ground & R-Ground & S-Ground & H-Verify & C-Verify \\ 
        \midrule
         \textcolor{gray}{\textit{Open-source (general-purpose)}} \\
         VideoChat2~\cite{li2024mvbench} & Vicuna-7B & 16 
        & 7.2 & 5.4 (74.5) & 1.0 (13.7) & 3.8 (52.1) & 3.6 (50.0) & 10.5 & 8.7 (82.8) & 0.6 (6.0) & 5.6 (53.3) & 5.4 (51.9) \\

        Video-LLaVA~\cite{lin2023videollava} & Vicuna-7B & 8 &
        9.4 & 7.6 (80.8) & 2.8 (30.3) & 5.0 (52.8) & 4.7 (50.0) 
        & \underline{13.4} & \underline{10.0} (74.5) & 3.1 (23.0) & 6.3 (46.9) & 7.0 (52.3) \\

        Video-LLaMA~\cite{videollama} & Vicuna-7B & 8 & 14.2 & 10.6 (74.9) & 5.3 (37.6) & 7.5 (53.3) & 7.3 (51.7) &12.8 & 8.5 (66.8) & \underline{7.2} (56.8) & \underline{7.3} (57.5) & \underline{7.5} (58.9) \\
        
        Video-ChatGPT~\cite{Maaz2023VideoChatGPT} & Vicuna-7B & 100 &
        14.4 & 12.8 (89.2) & 1.3 (8.8) & 6.5 (44.8) & 7.2 (50.0) & 3.3 & 2.8 (84.0) & 0.1 (4.1) & 1.7 (50.4) & 1.6 (49.2) \\

        Video-LLaMA2~\cite{cheng2024videollama2} & Mistral-7B & 8 
        & 20.0 & 16.8 (83.8) & 3.8 (19.0) & 10.3 (51.5) & 10.6 (52.9) & 10.4 & 8.2 (78.6) & 1.5 (14.8) & 5.4 (52.4) & 5.7 (54.7) \\
        \midrule
        
         \textcolor{gray}{\textit{Open-source (time-aware)}} \\
         VTG-LLM~\cite{guo2024vtgllm} & Llama2-7B & 96 
        & 26.0 & 16.1 (62.1) & \textbf{8.3} (32.0) & 6.0 (23.1) & 10.0 (38.4) & 6.8 & 5.3 (78.0) & 0.2 (3.0) & 0.7 (10.9) & 1.7 (24.9) \\

        VTimeLLM~\cite{vtimellm} & Vicuna-7B & 100 
        & \underline{27.3} & \underline{22.7} (83.2) & \underline{7.3} (26.9) & \underline{11.9} (43.7) & \underline{13.6} (49.8) & \textbf{31.9} & \textbf{26.4} (82.7) & \textbf{8.9} (27.9) & \textbf{14.1} (44.2) & \textbf{15.8} (49.5) \\

        TimeChat~\cite{timechat} & Llama2-7B & 96 
        & \bf 30.5 & \textbf{25.0} (82.1) & 5.6 (18.5) & \textbf{14.0} (45.9) & \textbf{15.6} (51.2) & 4.6 & 2.9 (64.1) & 1.0 (21.2) & 2.1 (46.7) & 2.4 (52.2) \\
        \midrule
        
         \textcolor{gray}{\textit{Closed-source}} \\
        \rowcolor{lightgray}
        GPT-4o~\cite{achiam2023gpt4} & - & 10 
        &  {28.5} &  {21.2 (74.3)} &  {9.3 (32.8)} &  {17.8 (62.4)} &  {20.3 (71.3)} &  {26.8} &  {18.1 (67.5)} &  {10.4 (38.8)} &  {16.5 (61.7)} &  {18.4 (68.8)} \\
        \rowcolor{lightgray}
        Gemini 1.5 Flash~\cite{reid2024gemini1.5} & - & 1 fps
        &  {34.6} &  {29.7 (85.7)} &  {24.8 (71.7)} &  {22.8 (65.8)} &  {24.5 (70.8)} &  {37.8} &  {30.8 (81.4)} &  {24.8 (65.6)} &  {22.4 (59.3)} &  {26.8 (70.8)} \\
      \bottomrule
      \end{tabular}
      }
    \caption{Consistency evaluation of \vllm~and closed-source models. The \textit{time-aware} models are specifically designed to grasp temporal moments. For each model, we specify the language model backbone (LLM) and the number of input frames (\# Frames) used. Relative consistency scores are in brackets. \vllm~often struggle to consistently respond to the probes, revealing their deficiencies in reliable video temporal understanding. In contrast, the closed-source models demonstrate relatively superior consistency across all probes.}
    \vspace{-3mm}
    \label{tbl: consistency}
\end{table*}

%% file: secs/4.Experiments.tex
\section{Experiments}
\label{sec: experiments}
In Table~\ref{tbl: consistency}, we present the evaluation results with 10 prominent models. The following sections introduce the tested models (\S\ref{sec: tested model introduction}) and analyze the results (\S\ref{sec:grounding_analysis} and \S\ref{sec:consistenty_analysis}).

\subsection{Tested Models}
\label{sec: tested model introduction}
\noindent\textbf{Open-source Models.} We select eight \vllm~whose official checkpoints are available, covering general-purpose conversation models: Video-LLaMA~\cite{videollama}, Video-LLaVA~\cite{lin2023videollava}, Video-ChatGPT~\cite{Maaz2023VideoChatGPT}, VideoChat2~\cite{li2024mvbench}, Video-LLaMA2~\cite{cheng2024videollama2} and models which are expressly time-aware: TimeChat~\cite{timechat}, VTimeLLM~\cite{vtimellm}, and VTG-LLM~\cite{guo2024vtgllm}. Typically, models like Video-LLaMA and Video-ChatGPT - we refer to these as general-purpose models - are developed by answering queries related to the video content. On the other hand, models like TimeChat are specifically designed to answer questions requiring precise timestamps within videos. We categorize these as time-aware models. The split allows us to assess how well the time-aware models have enhanced their grounding capabilities compared to the general-purpose ones and whether these advancements extend to consistency. 

\noindent\textbf{Closed-source Models.} We select GPT-4o~\cite{achiam2023gpt4} and Gemini 1.5 Flash~\cite{reid2024gemini1.5}, as they represent state-of-the-art closed-source models. Although they may not be specifically designed for temporal grounding, evaluating them allows us to explore how they handle such tasks.

\noindent\textbf{Implementation Details.} Since the closed-source and general-purpose models are not specifically designed for temporal grounding, we design prompts to guide them to include timestamps in their answers for reliable evaluation. Details on the models and their evaluation setup can be found in Appendix~\ref{appendix: explanation of vllms} and \ref{appendix: evaluation setup}.

\subsection{Grounding performance}
\label{sec:grounding_analysis}
Table~\ref{tbl: consistency} shows that the grounding performance varies across the models and datasets. The time-aware models demonstrate their superior grounding ability on Charades-CON, consistently outperforming the general-purpose models. While VTimeLLM performs well on both datasets, VTG-LLM and TimeChat fall short on ActivityNet-CON, showing significantly lower performance than on Charades-CON and underperforming most general-purpose models. This suggests their limitations for longer videos. The two closed-source models outperform most open-source models on both datasets, demonstrating their strong capabilities. 

\begin{figure*}[t]
        \centering
        \includegraphics[width=0.99\linewidth]{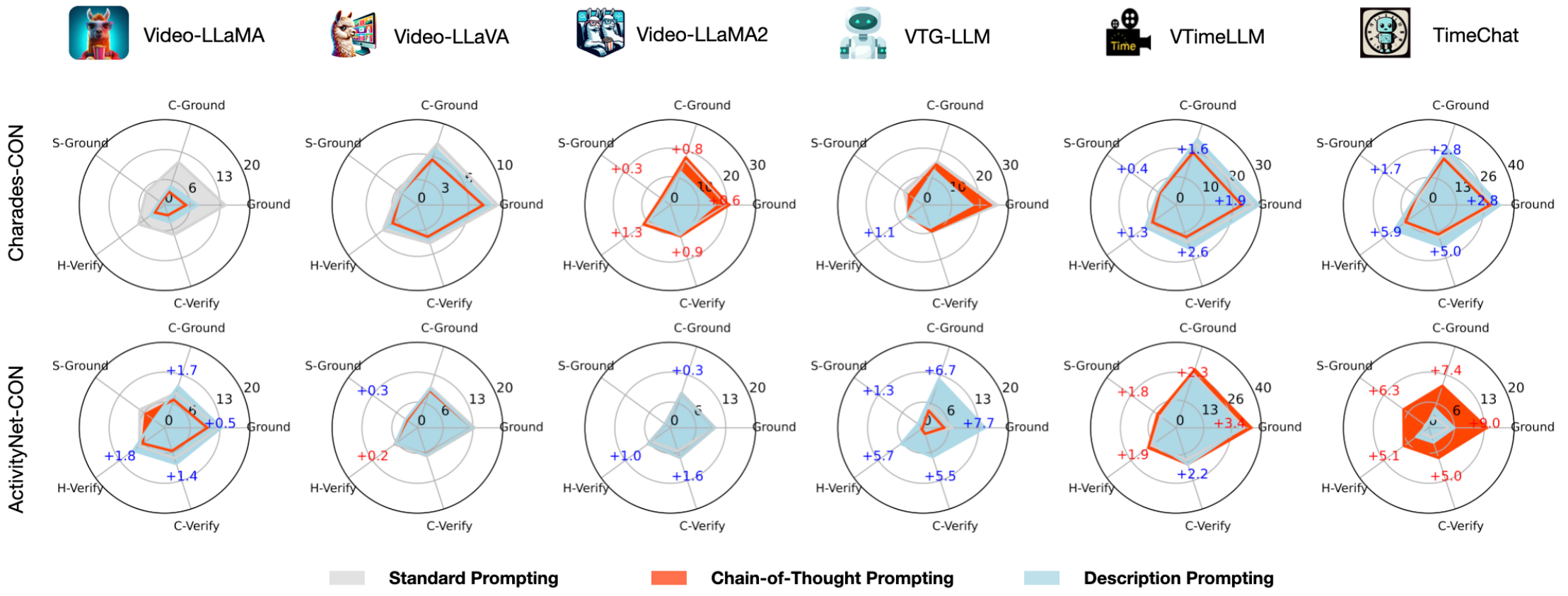}
    \caption{
    Consistency evaluation of \vllm~using different prompting methods. The Standard indicates the original performance. The highest improvement is highlighted in \textcolor{red}{red} for Chain-of-Thought prompting and in \textcolor{blue}{blue} for Description prompting.
    }
    \label{fig: prompt results}
    \vspace{-3mm}
\end{figure*}

\subsection{Consistency Analysis} 
\label{sec:consistenty_analysis}
\noindent\textbf{Rephrased grounding.} All models behave relatively well in this probe, with a relative consistency score exceeding 60\%, likely due to the use of LLMs for language comprehension. Yet, despite the variations in performance across different models, further investigations (refer to Appendix Table~\ref{tbl: mIoU of rephrased grounding}) reveal a common behavior -- models handle word replacement better than other modifications (\ie, active to passive conversation and word order modification). This suggests that further improvements are needed to address more complex structural variants for consistent grounding with rephrased queries.

\noindent\textbf{Shifted Grounding.} All open-source models struggle to adapt their predictions to shifted moments, consistently showing lower scores than in rephrased grounding. This suggests that they are less sensitive to visual variants than language variants. To further analyze this, we assess how differently the models predict timestamps for the original and shifted videos using the same query by measuring average IoU values between them. Interestingly, most models exhibit high IoU values in this comparison, reflecting their low performance in shifted grounding. Specifically, VideoChat2 and Video-LLaVA, with shifted grounding scores of 1.0 and 2.8 on Charades-CON, show high IoU values of 0.77 and 0.8. Although VTG-LLM shows the lowest IoU value among the open-source models at 0.53, Gemini achieves an even lower IoU value of 0.36 with superior performance in shifted grounding, solidifying its position as a leading model. Overall, these results suggest that most \vllm~may rely on language priors in queries for temporal grounding rather than on true comprehension of visual information~\cite{yuan2021closer, otani2020uncovering, xiao2024can}.

\noindent\textbf{Holistic Verification.} All open-source models demonstrate near chance-level verification consistency (with relative consistency scores around 50\%). Specifically, most models are only able to respond consistently for about half of their correct moment predictions, indicating significant room for improvement in their verification consistency. Interestingly, the time-aware models do not meet expectations here and fail to extend their grounding capabilities to verification consistency. Specifically, while VTimeLLM outperforms Video-LLaMA by 7.3\% in the main task on Charades-CON, this advantage shrinks to only 1.6\% in holistic verification, suggesting limited gains in consistency despite strong grounding performance. Notably, VTG-LLM shows significant deficiencies in providing consistent responses across verification probes. We speculate that perhaps the time-aware models are overly specialized in predicting timestamps while sacrificing QA capability. Please refer to Appendix~\ref{appendix: limitations of the time-aware models} for further discussion. Again, both closed-source models demonstrate strong verification consistency, significantly outperforming the open-source models. 

\noindent\textbf{Compositional Verification.} Most models perform better in compositional verification than in holistic verification. We conjecture that while the models may identify individual components of the video moment, they struggle to confirm the complete alignment of the query with the video moment. For better interpretation, we visualize some model responses to verification probes in Figure~\ref{fig: visualization}. TimeChat fails to verify its initial predictions based on the query but accurately confirms specific sub-information within it. Conversely, while Video-ChatGPT successfully verifies the query with its prediction, it struggles to identify the misaligned query. These inconsistencies between the probes highlight significant limitations in reliable video understanding.

%% file: secs/5.Solution_and_Discussion.tex
\input{tables/instruction-tuning}
\section{Solution and Discussion}
\label{sec: discussion}
This section explores potential solutions to improve consistency for \vllm. We first examine the impact of common prompting (\S\ref{sec: prompts}) and instruction tuning methods with target datasets on consistency (\S\ref{sec: instruction tuning}) and analyze their limitations. We then introduce \method, a targeted extension of instruction tuning to enhance consistency and grounding and demonstrate its effectiveness (\S\ref{sec: vtune}).

\subsection{Prompt Design for \vllm}
\label{sec: prompts}
We devise two prompting methods. The first method, inspired by Chain-of-Thought~\cite{wei2022chain} (CoT), provides step-by-step guidance to solve the target tasks. We describe each probe and offer instructions for tackling it sequentially. Specifically, for temporal grounding, we instruct the model first to find the key elements in the given query and analyze video features to identify the relevant moment. Our second method, description prompting (Desc), prompts the model to describe the video content before attempting the target tasks. We expect that this method encourages the model to provide answers that are more grounded in their video comprehension. Specifically, we prompt the model as ``\textit{Please describe the video in detail.}'' and then perform other tasks. Please refer to Appendix~\ref{appendix: prompting methods} for further details.

Figure~\ref{fig: prompt results} shows the results of \vllm~across the prompting methods. CoT demonstrates over a 5\% improvement across all aspects of TimeChat on ActivityNet-CON. Similarly, Desc improves all aspects of VTimeLLM and TimeChat on Charades-CON, as well as Video-ChatGPT and VTG-LLM on ActivityNet-CON. However, none of the prompting methods improve Video-LLaMA and Video-LLaVA on Charades-CON; in fact, they even degrade the performance. We also find that the prompting methods are ineffective for VideoChat2 (please refer to Appendix Figure~\ref{fig: prompt results for Video-ChatGPT and VideoChat2}). This is likely due to differences in model architecture and instruction tuning, \ie, whether the models are designed for such tasks. Overall, the improvements are often unstable, underscoring the limitations of prompting alone in enhancing temporal comprehension.
\begin{figure}[t]
        \centering
        \includegraphics[width=0.99\linewidth]{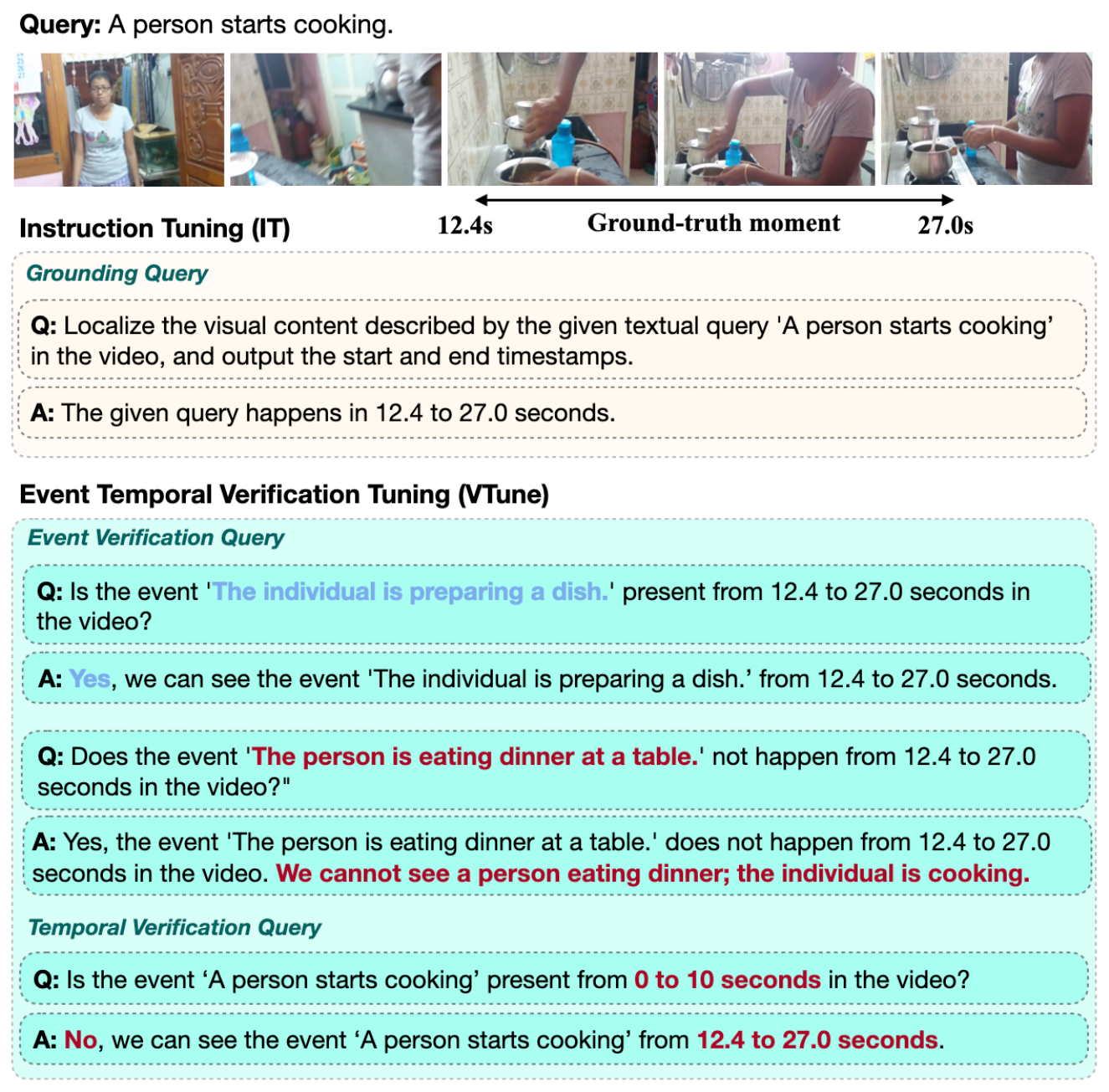}
    \vspace{-2mm}
    \caption{
    Visualization of instruction tuning methods. The \textcolor{blue}{blue} text represents content aligned with the meaning of original content, while the \textcolor{red}{red} text indicates irrelevant content. These colors also apply to the corresponding responses. While IT only requires a timestamp for the given query, \method~prompts the model to recognize temporal and content changes and respond with corrections.
    }
    \label{fig: instruction tuning}
    \vspace{-6.5mm}
\end{figure}

\subsection{Instruction Tuning on Target Datasets}
\label{sec: instruction tuning}
Instruction tuning (IT) in LLMs utilizes task-specific templates, generating the model's answer in a specific format. For temporal grounding, previous \vllm~\cite{vtimellm, timechat, qian2024momentor, qian2024momentor, guo2024vtgllm} use pre-defined templates like ``The given event occurs from \{start\} to \{end\} seconds.'' to include timestamps in their answer. To investigate the effect of instruction tuning on consistency, we select one general-purpose model, Video-LLaMA, and one time-aware model, TimeChat.

In Table~\ref{tbl: instruction tuning}, IT consistently boosts the model's grounding performance but remains ineffective in improving consistency. Specifically, although IT significantly improves the grounding performance of TimeChat by 25.3\% on Charades-CON, only a marginal improvement of 2.7\% is observed in holistic verification due to its low relative consistency score. Furthermore, it decreases the consistency of Video-LLaMA for shifted grounding. Likely, instruction tuning does not account for consistency. Specifically, the training objectives in LLMs, which focus on maximizing text token likelihoods, do not directly optimize consistency. In conclusion, while \vllm~perform well on traditional metrics and excel at targeted tasks, their predictions are not necessarily consistent or reliable. 
This motivates us to develop a new method to improve consistent temporal comprehension.
\subsection{Event Temporal Verification Tuning}
\input{tables/fine_tuned_results}
\input{tables/ablation}
\label{sec: vtune}
We introduce a simple yet effective method called Event Temporal Verification Tuning (\method). \method~aims to enhance the model's ability to distinguish between correct and incorrect vision-text correspondence between language queries and temporal moments, rather than naively instructing models to solve specific downstream tasks. Beyond predicting the timestamps of a given query, we tune the model to identify and correct inconsistencies in verification questions. 

As shown in Figure~\ref{fig: instruction tuning}, if we intentionally change the information in the query (\eg, from ``cooking'' to ``eating dinner''), the model should recognize the changes by comparing the query with the actual video content, and correctly respond with reasoning. We refer to these queries as verification queries and prompt GPT-4o mini to generate them with corresponding corrections. Additionally, we pair queries with non-ground-truth moments to encourage the model to relocate to the right moment accurately (\eg, from ``0 to 10 seconds'' to ``12.4 to 27.0 seconds''). We refer to these queries as temporal verification queries. Further details of \method~can be found in Appendix~\ref{appendix: instruction tuning}.

\begin{figure}[t]
        \centering
        \includegraphics[width=0.99\linewidth]{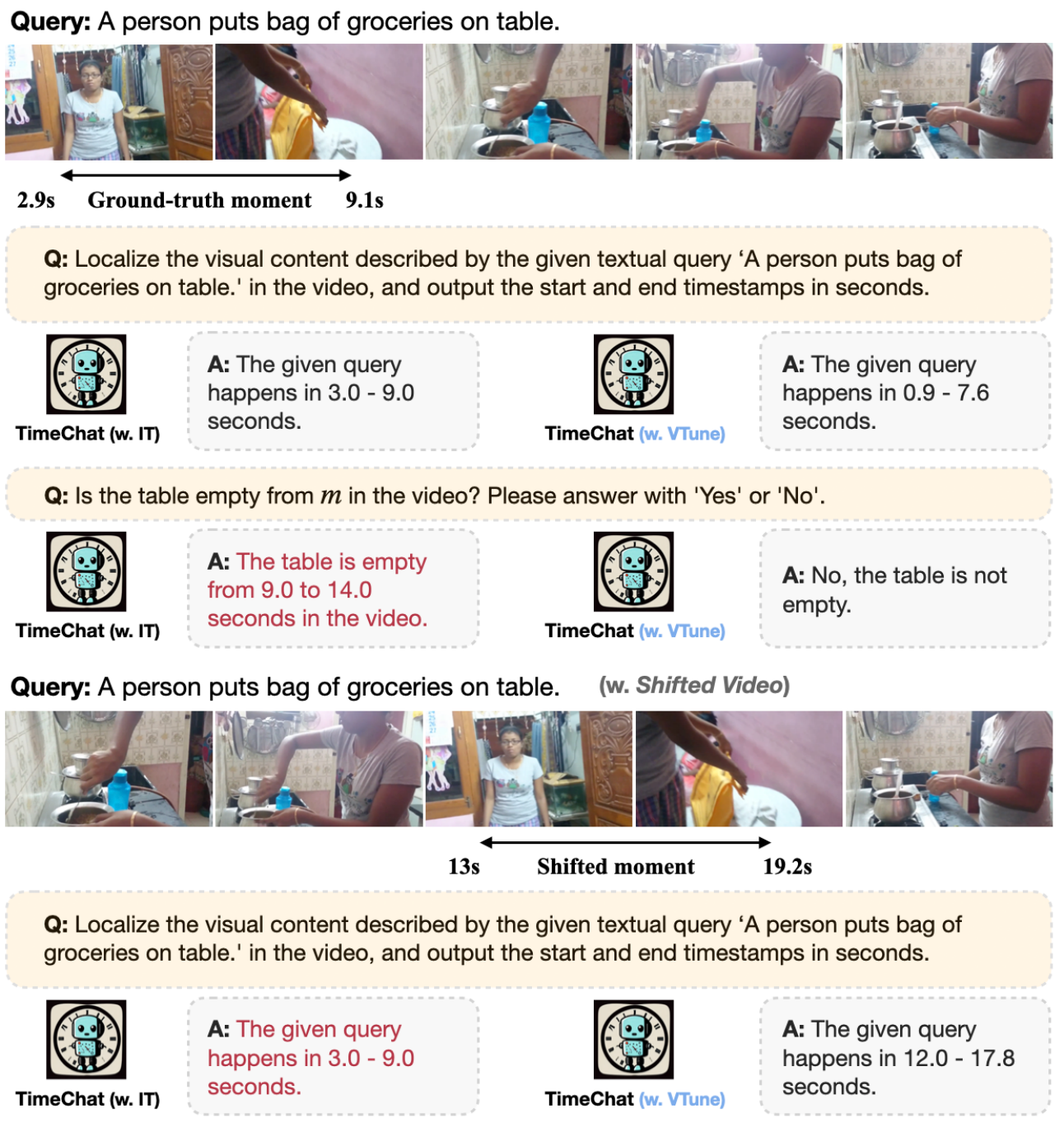}
        \vspace{-3mm}
    \caption{
     Responses of the instruction-tuned models. While IT fails to verify the compositional query and adjust its predictions to the shifted moment, \method~enables the model's consistent responses.
    }
    \label{fig: responses of instruction-tuned models}
    \vspace{-4mm}
\end{figure}

In Table~\ref{tbl: instruction tuning}, \method~demonstrates significant improvements in both grounding and consistency, unlike the unstable improvements observed with the previous solutions. Notably, the improved consistency achieved via VTune brings substantial gains in grounding performance, significantly outperforming IT. Additionally, we provide fine-tuned performance on the original test sets in Table~\ref{tbl: fine-tuned results}. \method~continues to demonstrate positive effects, with TimeChat achieving state-of-the-art results on both datasets. To further analyze on \method, we provide an ablation study of instruction tuning in Table~\ref{tbl: ablation of instruction tuning}. Both models display consistent improvements in grounding and consistency when tuned with event verification queries, with further gains from the addition of temporal verification queries. Figure~\ref{fig: responses of instruction-tuned models} qualitatively support these results. The above results demonstrate \method' effectiveness in robust and consistent temporal comprehension.

%% file: tables/instruction-tuning.tex
\begin{table*}[t]
    \resizebox{\linewidth}{!}{
    \begin{tabular}{lccccccccccc}
    \toprule
    \multirow{2}{*}{\bf Method} & \multirow{2}{*}{\bf FT} & \multicolumn{5}{c}{\bf Charades-CON} & \multicolumn{5}{c}{\bf ActivityNet-CON} \\
    \cmidrule(lr){3-7} \cmidrule(lr){8-12}
    & & Ground & R-Ground & S-Ground & H-Verify & C-Verify 
    & Ground & R-Ground & S-Ground & H-Verify & C-Verify \\
    \midrule
    \rowcolor{lightgray}
    Video-LLaMA~\cite{videollama} & \xmark & 14.2 & 10.6 (74.9) & 5.3 (37.6) & 7.5 (53.3) & 7.3 (51.7) & 12.8 & 8.5 (66.8) & \underline{7.2} (56.8) & 7.3 (57.5) & 7.5 (58.9) \\
    
    Video-LLaMA {\small (w. CoT)} & \xmark
    & \textcolor{red}{5.0} & \textcolor{red}{3.3} (66.6) & \textcolor{red}{1.1} (22.2) & \textcolor{red}{2.9} (58.7) & \textcolor{red}{2.5} (50.0) 
    & \textcolor{red}{10.2} & \textcolor{red}{6.9} (68.0) & \textcolor{red}{5.5} (54.4) & \textcolor{red}{6.2} (61.5) & \textcolor{red}{5.7} (56.2) \\
    
    Video-LLaMA {\small (w. Desc)} & \xmark
    & \textcolor{red}{7.7} & \textcolor{red}{4.7} (60.6) & \textcolor{red}{1.5} (20.0) & \textcolor{red}{4.3} (55.5) & \textcolor{red}{4.2} (55.0) 
    & \textcolor{blue}{13.3} & \textcolor{blue}{10.2} (73.3) & \textcolor{red}{3.1} (22.2) & \textcolor{blue}{9.1} (65.3) & \textcolor{blue}{8.9} (64.1) \\
    
    Video-LLaMA {\small (w. IT)} & \cmark 
    & \textcolor{blue}{\underline{45.1}} & \textcolor{blue}{\underline{32.0}} (71.8) & \textcolor{blue}{\underline{8.9}} (19.7) & \textcolor{blue}{\underline{24.8}} (54.9) & \textcolor{blue}{\underline{23.4}} (51.8)  
    & \textcolor{blue}{\underline{20.6}} & \textcolor{blue}{\underline{16.4}} (80.7) & \textcolor{red}{6.5} (32.0) & \textcolor{blue}{\underline{11.7}} (57.3) & \textcolor{blue}{\underline{11.1}} (54.1) \\
    
    Video-LLaMA {\small \bf (w. \method)} & \cmark 
    & \textcolor{blue}{\bf54.4} & \textcolor{blue}{\bf38.2} (70.3) & \textcolor{blue}{\bf10.9} (20.0)  & \textcolor{blue}{\bf30.7} (56.5) & \textcolor{blue}{\bf30.0} (55.2)
    & \textcolor{blue}{\bf33.0} & \textcolor{blue}{\bf24.7} (74.8) & \textcolor{blue}{\bf10.0} (30.2) & \textcolor{blue}{\bf20.2} (61.1) & \textcolor{blue}{\bf17.7} (53.7) \\
    \midrule
    \rowcolor{lightgray}
    TimeChat~\cite{timechat} & \xmark
    & 30.5 & 25.0 (82.1) & 5.6 (18.5) & 14.0 (45.9) & 15.6 (51.2)
    & 4.6 & 2.9 (64.1) & 1.0 (21.2) & 2.1 (46.7) & 2.4 (52.2) \\
    
    TimeChat {\small (w. CoT)} & \xmark
    & \textcolor{red}{28.7} & \textcolor{red}{22.8} (79.4) & \textcolor{blue}{7.1} (24.6) & \textcolor{red}{13.5} (46.9) & \textcolor{red}{14.4} (50.2)
    & \textcolor{blue}{13.6} & \textcolor{blue}{10.3} (75.9) & \textcolor{blue}{7.3} (53.6) & \textcolor{blue}{7.2} (52.8) & \textcolor{blue}{7.4} (54.4) \\
    
    TimeChat {\small (w. Desc)} & \xmark
    & \textcolor{blue}{33.3} & \textcolor{blue}{27.8} (83.4) & \textcolor{blue}{7.3} (22.0) & \textcolor{blue}{19.9} (59.9) & \textcolor{blue}{20.6} (61.8)
    & \textcolor{blue}{5.9} & \textcolor{blue}{4.7} (79.7) & \textcolor{red}{0.7} (11.9) & \textcolor{blue}{3.6} (60.3) & \textcolor{blue}{3.6} (61.4) \\
    
    TimeChat {\small (w. IT)} & \cmark 
    & \underline{\textcolor{blue}{55.8}} & \underline{\textcolor{blue}{50.9}} (91.3) & \underline{\textcolor{blue}{10.5}} (18.9) & \underline{\textcolor{blue}{16.7}} (30.0) & \underline{\textcolor{blue}{25.7}} (46.2)
    & \underline{\textcolor{blue}{25.3}} & \underline{\textcolor{blue}{20.2}} (80.4) & \underline{\textcolor{blue}{7.5}} (29.9) & \underline{\textcolor{blue}{8.7}} (34.5) & \underline{\textcolor{blue}{12.6}} (49.9) \\

    TimeChat {\small \bf (w. \method)} & \cmark 
    & \textcolor{blue}{\bf 76.2} & \textcolor{blue}{\bf69.2} (90.8) & \textcolor{blue}{\bf36.2} (47.5) & \textcolor{blue}{\bf44.8} (58.8) & \textcolor{blue}{\bf42.4} (55.7)
    & \textcolor{blue}{\bf37.4} & \textcolor{blue}{\bf28.3} (75.6) & \textcolor{blue}{\bf10.6} (28.3) & \textcolor{blue}{\bf19.6} (52.3) & \textcolor{blue}{\bf19.3} (51.5) \\
    \bottomrule
    \end{tabular}
    }
    \caption{Consistency evaluation of \vllm~with the proposed solutions. Performance gains are highlighted in \textcolor{blue}{blue}, while decreased performances are shown in \textcolor{red}{red}. FT represents whether the model is fine-tuned. Compared to the previous solutions (\ie CoT, Desc, and IT), \method~demonstrates significant improvements for both grounding and consistency across the models and datasets.}
    \label{tbl: instruction tuning}
    \vspace{-3mm}
\end{table*}

%% file: tables/fine_tuned_results.tex
\begin{table}[t]
    \resizebox{\linewidth}{!}{
      \begin{tabular}{lcccc}
      \toprule
      \multirow{2}{*}{\bf Method} & \multicolumn{2}{c}{\bf Charades-STA} & \multicolumn{2}{c}{\bf ActivityNet-Captions} \\
      \cmidrule{2-3} \cmidrule{4-5}
       & R@1, 0.5 & R@1, 0.7 & R@1, 0.5 & R@1, 0.7 \\ \midrule
       \textcolor{gray}{\textit{Task-Specific Models}} \\
       BM-DETR~\cite{Jung_2025_WACV} & \textcolor{gray}{59.4} & \textcolor{gray}{38.3} 
       & \textcolor{gray}{49.6} & \textcolor{gray}{30.6} \\ 
       Mr.BLIP~\cite{boris2024mrblip} & \textcolor{gray}{69.3} & \textcolor{gray}{49.2} & \textcolor{gray}{53.9} & \textcolor{gray}{35.5} \\ \midrule
       
       \textcolor{gray}{\textit{\vllm}} \\
       HawkEye~\cite{wang2024hawkeye} & \underline{58.3} & 28.8 & \underline{34.7} & 17.7 \\
       VTG-LLM~\cite{guo2024vtgllm} & 57.2 & \underline{33.4} & - & - \\ \midrule 
       
       Video-LLaMA~\cite{videollama} & 35.0 & 18.6 & 25.2 & 14.4 \\
       \bf Video-LLaMA-VT (Ours) & 37.1 & 20.1 & 34.3 & \underline{19.1} \\ \midrule
       
       TimeChat~\cite{timechat} & 46.7 & 23.7 & 28.0 & 15.8 \\
       \bf TimeChat-VT (Ours) & \textbf{58.4} & \textbf{34.7} & \bf 41.0 & \bf 23.7 \\
      \bottomrule
      \end{tabular}
      }
    \caption{
    Fine-tuned performances on the original test sets. Task-specific models are designed to perform only a single target task. \method~effectively improves grounding performance in both models.
    }
    \label{tbl: fine-tuned results}
\end{table}

%% file: tables/ablation.tex
\begin{table}[t]
    \resizebox{\linewidth}{!}{
      \begin{tabular}{lcccccccccc}
      \toprule
      \multirow{2}{*}{\bf Method} & \multirow{2}{*}{\bf G} & \multirow{2}{*}{\bf E} & \multirow{2}{*}{\bf T} & \multicolumn{5}{c}{\bf Charades-CON} \\
      \cmidrule(lr){5-10}
      & & & & Ground & Ground (0.7) & R-Ground & S-Ground & H-Verify & C-Verify \\
       \midrule
       \multirow{3}{*}{Video-LLaMA}       
       & \cmark & &
       & 45.1 & 23.3 & 32.0 (71.8) & 8.9 (19.7) & 24.8 (54.9) & 23.4 (51.8)  \\
        
        & \cmark & \cmark & 
        & \underline{48.8} & \underline{27.8} & 35.1 (71.9) & 10.2 (20.8) & \underline{28.0} (57.4) & \underline{26.4} (54.1) \\
       
       & \cmark & \cmark & \cmark 
       & \bf 54.4 & \bf 36.6 & \textbf{38.2} (70.3) & \textbf{10.9} (20.0) & \textbf{30.7} (56.5) & \textbf{30.0} (55.2) \\       
       \midrule
       
       \multirow{3}{*}{TimeChat}        
       & \cmark & &
       & 55.8 & 30.2 & 50.9 (91.3) & 10.5 (18.9) & 16.7 (30.0) & 25.7 (46.2) \\

       & \cmark & \cmark & 
       & \bf 76.2 & \underline{52.3} & \underline{68.7} (90.1) & \underline{15.5} (20.4) & \underline{40.4} (53.0) & \underline{40.5} (53.1) \\

       & \cmark & \cmark & \cmark 
       & \bf 76.2 & \bf 58.8 & \textbf{69.2} (90.8) & \textbf{36.2} (47.5) & \textbf{44.8} (58.8) & \textbf{42.4} (55.7) \\
      \bottomrule
      \end{tabular}
      }
    \caption{Query effectiveness by type. G represents grounding queries, and E and T indicate event and temporal verification queries. Additionally, we report ``R@1, IoU=0.7'' in Ground (0.7).
    }
    \label{tbl: ablation of instruction tuning}
    \vspace{-4mm}
\end{table}

%% file: secs/6.Conclusion.tex
\section{Conclusion}
We have studied the consistency of \vllm~in temporal comprehension. We propose dedicated evaluation datasets and a series of dedicated probes and find that most \vllm~exhibit inconsistent behaviors, unveiling their significant deficiencies in reliable video comprehension. We further demonstrate the limitations of common prompting methods and instruction tuning with target datasets, revealing their unstable improvements. To this end, we propose~\method, which explicitly accounts for consistency, demonstrating its significant improvement in both grounding and consistency. We hope that future \vllm~will consider consistency towards trustworthiness, in addition to standard tasks.

\section*{Acknowledgements}
This research was funded by the NUS Artificial Intelligence Institute (NAII) seed grant number NAII-SF-2024-003 and was partly supported by the IITP (RS-2021-II212068-AIHub/10\%, RS-2021-II211343-GSAI/15\%, RS-2022-II220951-LBA/15\%, RS-2022-II220953-PICA/20\%), NRF (RS-2024-00353991-SPARC/20\%, RS-2023-00274280-HEI/10\%), and KEIT (RS-2024-00423940/10\%) grant funded by the Korean government.

%% file: secs/appendix.tex
\clearpage
\setcounter{section}{0}
\appendix

In this file, we provide additional details as follows:

\begin{enumerate}[label=\Alph*.]
    \item \textbf{Details of Evaluation Datasets}: We provide the details of our proposed datasets, Charades-CON and ActivityNet-CON.

    \item \textbf{Details of Models}: We describe the \vllm~that are utilized in our experiments one by one. Then we provide the experiment setup of the models for our evaluation. Additionally, we further discuss the limitations of the time-aware models.

    \item \textbf{Details of Prompting Methods}: We describe two prompting methods, including chain-of-thought and description prompting.
    
    \item \textbf{Details of Instruction Tuning}: We detail the process of instruction tuning and \method.

    \item \textbf{Additional Experiments}: We provide additional experiments to provide further insights.

    \item \textbf{Prompt Designs}: In this study, we design various prompts to construct evaluation sets, evaluate the model's responses, and generate verification queries. We provide detailed prompts for each.
\end{enumerate}

\section{Details of Datasets}
In Table~\ref{tbl: dataset_details}, we provide statistics of each dataset. The length of the video and their moments remain almost the same compared to the original datasets. Generated aligned and misaligned queries have similar lengths compared to the ground-truth queries. 

\section{Details of \vllm}
\label{appendix: model details}
\subsection{\vllm}
\label{appendix: explanation of vllms}
In this section, we describe eight open-source \vllm: Video-LLaMA~\cite{videollama}, Video-LLaVA~\cite{lin2023videollava}, Video-ChatGPT~\cite{Maaz2023VideoChatGPT}, VideoChat2~\cite{li2024mvbench}, Video-LLaMA2~\cite{cheng2024videollama2}, TimeChat~\cite{timechat}, VTimeLLM~\cite{vtimellm}, and VTG-LLM~\cite{guo2024vtgllm}, which are utilized in our evaluation. Note that we experiment with a single A100 GPU for all models.

\begin{enumerate}
\item \textbf{Video-LLaMA} exhibits two branches: Vision-Language and Audio-Language in its modeling and utilizes cross-modal training from both the frozen pre-trained visual and audio encoder. It shows a remarkable zero-shot audio understanding capability and also generates responses to the visual and audio information presented in the videos. We select the fine-tuned checkpoint: \textit{Video-LLaMA-2-7B-Finetuned}, which is additionally fine-tuned on the instruction tuning data from Mini-GPT-4~\cite{zhu2023minigpt}, LLaVA~\cite{liu2023llava}, and VideoChat~\cite{li2023videochat}.

\item \textbf{Video-LLaVA} enables an LLM to perform visual reasoning capabilities, and it leads to comprehending both images and videos simultaneously. To do this, Video-LLaVA collects both image and video-text datasets and incorporates them in its instruction tuning. 

\item \textbf{Video-ChatGPT} designs spatiotemporal video modeling and constructs video instruction tuning upon LLaVA~\cite{liu2023llava}. It introduces a new dataset for video instruction tuning, containing 100,000 high-quality video-instruction pairs. Video-ChatGPT outperforms previous \vllm ~in Zero-shot VQA across several benchmarks. Additionally, Video-ChatGPT proposes a video conversation evaluation framework. 
\begin{figure}[t]
        \centering
        \includegraphics[width=0.99\linewidth]{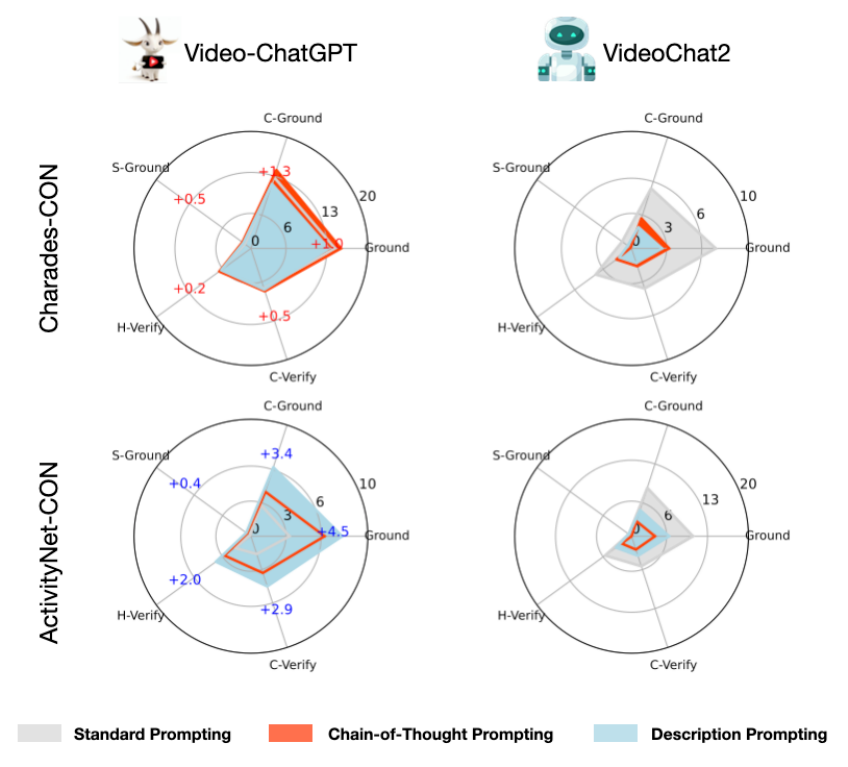}
    \caption{
    Consistency evaluation results of Video-ChatGPT and VideoChat2 using different prompting methods. While the prompting methods improve Video-ChatGPT, they are ineffective for VideoChat2.
    }
    \label{fig: prompt results for Video-ChatGPT and VideoChat2}
\end{figure}
\input{appendix_tables/dataset_details}
\input{appendix_tables/templates_for_holistic_verification}

\item \textbf{VideoChat2} design a progressive multimodal training. Specifically, they aim to align vision and text in the first stage and then generate captions from various image-text pairs. Finally, they conduct instruction tuning to better align responses with human instructions. VideoChat2 demonstrates significant improvements in Zero-Shot Video Question Answering, video conversation, and own built comprehensive video benchmarks. 

\item \textbf{Video-LLaMA2} is one of the state-of-the-art \vllm, demonstrating superior performances on video question-answering tasks. They seamlessly integrate both visual and audio modalities in videos and propose STC connector to understand spatiotemporal video information. 
\input{appendix_tables/grounding_templates}

\item \textbf{TimeChat} is specifically developed to accurately localize and understand specific video moments from long videos. TimeChat designs two key architectural: (1) A time-aware frame encoder that explicitly encodes video frames along with timestamps, (2) A sliding video Q-Former to accommodate sequential information in video frames. Specifically, TimeChat borrows the image Q-Former from InstructBLIP~\cite{dai2023instructblip} and the video Q-former from Video-LLaMA~\cite{videollama}. Additionally, TimeChat includes the instruction tuning datasets of Valley~\cite{luo2023valley} and constructs 125K video instruction tuning datasets to perform time-related video understanding tasks, such as VTG and DVC. 

\item \textbf{VTimeLLM} proposes a three-stage temporal-aware method, including image-text training and understanding events within the video, enabling more precise video temporal understanding. VTimeLLM devises two types of QA dialogue templates, including single-turn and multi-turn, to prompt questions requiring a comprehensive description of all events and their corresponding timestamps. 

\item \textbf{VTG-LLM} utilizes specific modules to improve temporal representations in Video-LLM. While previous \vllm~suffer quantization errors by using special time tokens, VTG-LLM use absolute-time tokens (\eg, \text{<TIME$\_$ZERO>, <TIME$\_$ONE>}) to resolve this and handle videos more than one hour. The model is instruction-tuned with VTG-IT-120K,  a high-quality and comprehensive instruction tuning for time-related video understanding tasks. 
\end{enumerate}
\input{appendix_tables/grounding_results}
\input{appendix_tables/grounding_GPT4o}

\subsection{Evaluation Setup}
\label{appendix: evaluation setup}
As previously mentioned in Section~\ref{sec: experiments}, the feasibility of general-purpose \vllm~and closed-source models for temporal grounding is absent. In this section, we provide adaptation details for these models. 

\noindent\textbf{General-purpose \vllm.} We empirically found that the general-purpose \vllm~are very sensitive to prompts, easily failing to extract timestamps from their answers. For instance, the model does not meet our expectations, such as answering ``I'm unable to find timestamps in the video'', or the model's answer only includes a single timestamp like ``The given event happens at 3 seconds.'' We categorize these answers as \textit{Invalid output}. Therefore, we carefully design several prompts for temporal grounding to find the stable setting for each model and provide experiment results in Table~\ref{tbl: finding the optimal setting for temporal grounding}. For Video-LLaMA, we have two options for LLM backbones, Llama-2 and Vicuna. While using LLaMA-2 as an LLM backbone and the first prompt achieves the best grounding performance, it demonstrates unstable performance depending on the prompt. Therefore, we utilize Vicuna and the first prompt (\ie, [1] in Table~\ref{tbl: grounding prompt templates}), demonstrating better consistency (see Table~\ref{tbl: consistency of vllm across different settings}). Other \vllm~like Video-ChatGPT are fixed with a specific LLM backbone, so we follow their development settings. 
In conclusion, we use the first prompt for Video-LLAMA and VideoChat2, the second prompt for Video-LLAVA, and the third prompt for Video-ChatGPT and VideoLLaMA2.
\begin{figure*}[t]
        \centering
        \includegraphics[width=0.99\linewidth]{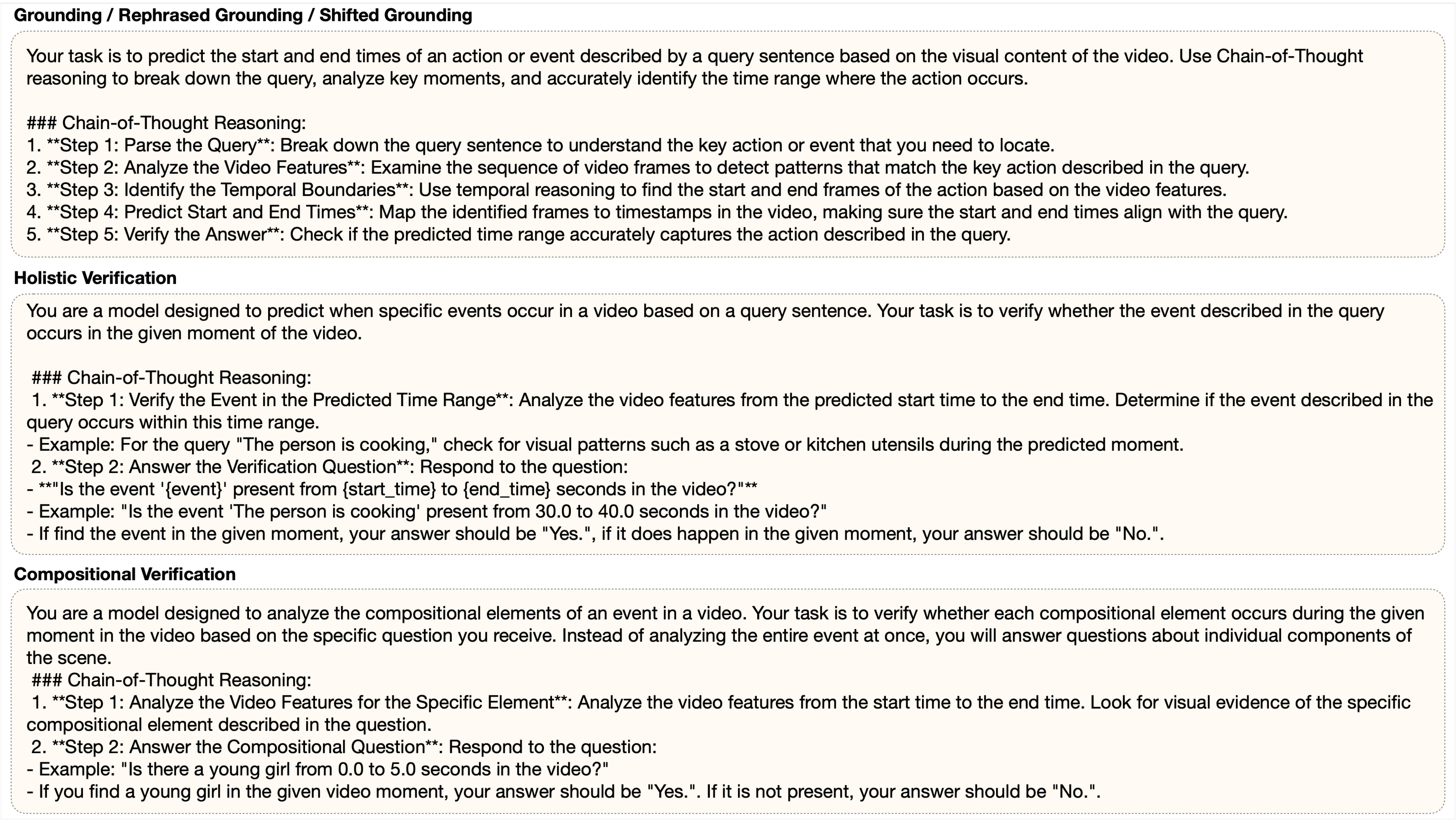}
    \caption{
    Chain-of-Thought prompts for temporal grounding and the probes. Since temporal grounding, rephrased grounding, and shifted grounding share the same goal, predicting the video moment accurately, they use the same prompt. For verification probes, we specify which parts the model should focus on, whether it’s the entire sentence or specific sub-elements.
    }
    \label{fig: COT prompts}
\end{figure*}

\begin{figure}[t]
        \centering
        \includegraphics[width=0.99\linewidth]{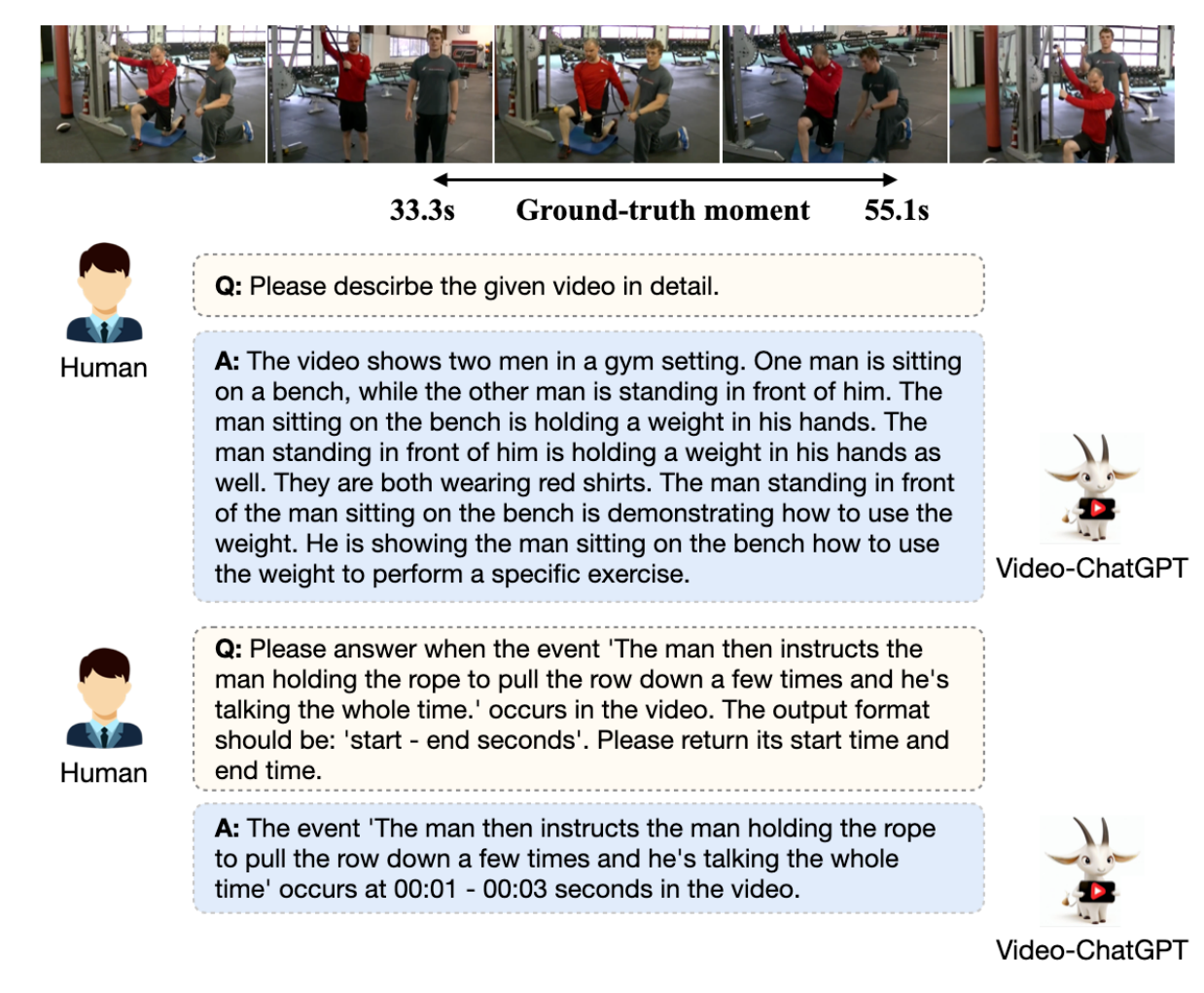}
    \caption{
    The responses of Video-ChatGPT with Description prompting. While Video-ChatGPT generates a rich video caption, it contains incorrect information that may mislead accurate predictions. In this case, Video-ChatGPT fails to accurately predict the timestamp in the video.
    }
    \label{fig: description prompts}
\end{figure}

\noindent\textbf{Time-aware \vllm.} For TimeChat, VTimeLLM, and VTG-LLM, we follow the same prompt templates, which are used in their official instruction tuning for temporal grounding. Likewise, we use the official codes to extract the timestamps from their predictions.

\noindent\textbf{Closed-source Models.} Given the considerable budget required to utilize GPT-4o, we first define the set of frame numbers \{10, 32, 64\} and sub-sample 100 videos from each dataset to efficiently find the optimal setting. In Table~\ref{tbl: GPT-4o grounding}, we provide the grounding performance of GPT-4o across different frame numbers. Note that the performances in Table~\ref{tbl: GPT-4o grounding} are different from those in Table~\ref{tbl: consistency} because the number of test samples is different. During the experiment, we instruct GPT-4o on how many frames are in the given video and how long each frame is sampled. We find that using 10 frames shows the best performance. For Gemini, we use the official APIs to process video inputs.

\noindent\textbf{Visualization.} We present example question-answer pairs for all models in Table~\ref{tbl: qa examples}. The format of timestamps varies depending on the model, so we design a timestamp extraction function for each model.

\subsection{Limitations of the Time-aware Models}
\label{appendix: limitations of the time-aware models}
We hypothesize that the excessive focus on temporal grounding with specialized elements in the time-aware models may limit its adaptability to various video understanding tasks and hinder generalization capabilities. For instance, TimeChat develops a time-aware frame encoder that explicitly encodes video features along with the corresponding sentences containing timestamps, such as ``This frame is sampled at 2s.'' VTG-LLM utilizes digit time tokens, such as \text{<TIME$\_$ZERO> and <TIME$\_$ONE>}, to effectively represent timestamps in videos. While such methods may be effective for answering questions that require timestamps within videos, providing these timestamp priors could hinder generalization across various video understanding tasks. This limitation is particularly evident in VTG-LLM, which shows poor relative consistency for verification probes, and we empirically find that VTG-LLM often struggles to follow instructions. Specifically, VTG-LLM frequently generates timestamps as answers, even when asked for verification probes.
\section{Details of Prompting Methods}
\label{appendix: prompting methods}
In this section, we provide details of the two prompting methods, Chain-of-Thought and Description prompting.

\noindent\textbf{Chain-of-Thought.} We provide prompts that we utilize for each prompting method. In Figure~\ref{fig: COT prompts}, we give three chain-of-thought prompts that we design for each probe in our evaluation. The three tasks, temporal grounding, rephrased grounding, and shifted grounding, share the same prompt since their output is the same. For holistic and compositional verification, we instruct on which part the model should concentrate either a whole sentence or sub-elements.

\noindent\textbf{Description Prompting.} We first ask the model ``Please describe the given video in detail.'' Then we keep the generated video description in chat history to perform the probes during our evaluation. We give a visual example of Video-ChatGPT using the description prompting method in Figure~\ref{fig: description prompts}. In our previous discussion, we highlighted that while using the description prompting method enhances the model's consistency, it may lead to negative outcomes such as generating incorrect descriptions or hallucinations. As illustrated in Figure~\ref{fig: description prompts}, Video-ChatGPT provides a detailed description of the video, but some of the information in the generated description is inaccurate. For instance, the video depicts two men in a gym, with only one of them wearing a red shirt. However, the model incorrectly states that both men are wearing red shirts. This might lead to inaccurate moment predictions.

\section{Details of Instruction Tuning}
\label{appendix: instruction tuning}
In this section, we provide a detailed experiment setup for instruction tuning and \method.

\noindent\textbf{Instruction Tuning.} Following the previous work~\cite{timechat}, we collect the annotations in the train split and convert the annotations into a task-specific template to derive \vllm~can predict the timestamps. For example, if the query ``The person closes the laptop.'' is grounded in 0 to 5 seconds in the video, we prompt the models ~``Localize the visual content described by the given textual query `The person closes the laptop.' in the video, and output the start and end timestamps in seconds.''. Then the model's answer should be ``The given query happens in 0 - 5 seconds.'' We utilize the official codes and configurations to conduct instruction tuning upon their official checkpoints. 
\input{appendix_tables/grounding_with_IoUs}

\begin{figure*}[t]
        \centering
        \includegraphics[width=0.99\linewidth]{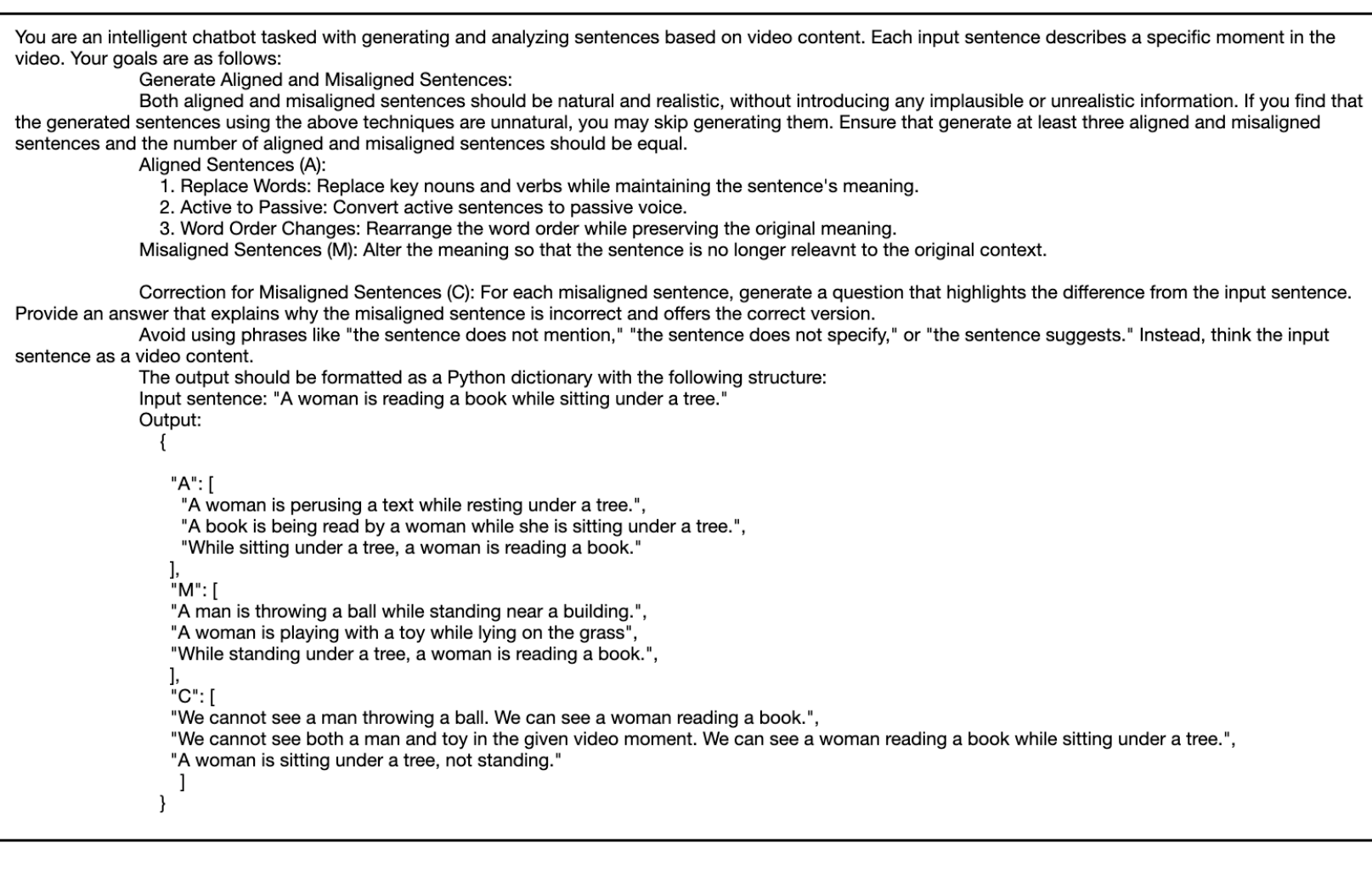}
    \caption{
    The prompt to generate verification questions for \method.
    }
    \label{fig: prompt for ETVT}
\end{figure*}

\noindent\textbf{Event Temporal Verification Tuning.} We construct verification questions from the training split of each dataset. Specifically, we generate both aligned and misaligned sentences from the original query and also generate correction information on what information is wrong in misaligned sentences by providing the prompt in Figure~\ref{fig: prompt for ETVT} to GPT-4o-mini. Then we pair aligned sentences with affirmative responses and misaligned sentences with denial responses and their corresponding corrections.

\noindent\textbf{Experiment Settings.} The number of queries by type for each dataset can be found in Table~\ref{tbl: the number of queries}. For \method, we follow the official configurations and only change hyper-parameters, including iters per epoch and warmup steps, to reflect the number of annotations for each training. We report hyper-parameters used in each instruction tuning method in Table~\ref{tbl: hyperparameters}. We use 4$\times$80 GB A100 GPUs for all experiments and use the same prompts and hyper-parameters in both Video-LLaMA and TimeChat. In our setting, Video-LLaMA requires less than one day, and TimeChat requires at least two days for instruction tuning. For \method, Video-LLaMA takes under two days, while TimeChat requires six days.
\input{appendix_tables/hyperparameters}
\input{appendix_tables/statistics_of_queries}
\input{appendix_tables/consistency_across_different_prompts}

\input{appendix_tables/rephrased_grounding}

\section{Additional Experiments}
\label{appendix: additional experiments}
In this section, we provide additional experiment results to further provide insights. We first provide the consistency evaluation results using different IoU values. Then we examine how the prompt for temporal grounding and scalability of LLM impact consistency. Then we further analyze the results of rephrased grounding and shifted grounding.

\noindent\textbf{Consistency Results across Different IoU thresholds.} In Table~\ref{tbl: grounding and consistency across different IoUs}, we provide additional experiment results using an additional IoU threshold (\ie, 0.7) for temporal grounding and grounding probes.

\noindent\textbf{The impact of Temporal Grounding Prompt on Consistency.} 
\label{appendix: impact of prompt and LLM scale}
To investigate this, we provide the results in Table~\ref{tbl: consistency of vllm across different settings}. As the prompt designs affect the grounding performance of \vllm, we also investigate their consistency. Specifically, we select a series of settings that demonstrate high performance for each model in Table~\ref{tbl: finding the optimal setting for temporal grounding}. While Video-LLaMA with Vicuna-7B and the second prompt demonstrate better relative consistency scores for verification probes, there is no significant performance gap compared to other settings. Likewise, for Video-LLaVA, Video-ChatGPT, and Video-LLaMA2, the grounding prompt does not significantly impact the model's performance.

\noindent\textbf{The impact of Scalability of LLM on Consistency.} Additionally, we evaluate Video-LLaMA with Vicuna-13B and confirm that there are no significant improvements. Specifically, while it shows higher relative consistency in rephrased grounding, it fails to demonstrate strong capabilities for other probes. Likely, a larger-scale LLM does not necessarily represent improved consistency. Unfortunately, we cannot evaluate the models with larger-scale LLM backbones, as most of them have either conducted experiments only with 7B LLMs or have not released the corresponding checkpoints. For instance, TimeChat and VTG-LLM haven't officially conducted experiments using 13B LLM backbones. VTimeLLM reports the performance using Vicuna-13B in the original paper, but has not officially released the corresponding checkpoint.

\noindent\textbf{Analysis of Rephrased Grounding.} In Table~\ref{tbl: mIoU of rephrased grounding}, we provide the performance for rephrased grounding by three modification rules. The rules are previously introduced in~\Cref{sec: dataset} including word replacement, active-to-passive conversion, and word order change. We confirm that most models temporally ground rephrased queries with word replacement. 
As we discussed in~\Cref{sec: experiments}, this suggests the need for improvements in consistent grounding and demonstrates the effectiveness of our modification rules providing various language variations.

\begin{figure}[t]
        \centering
        \includegraphics[width=0.99\linewidth]{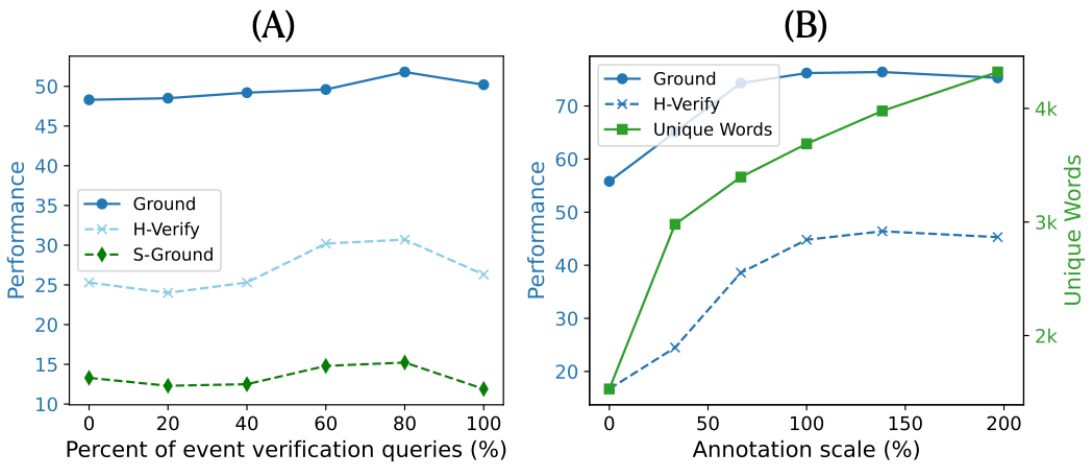}
    \caption{
    \textbf{Experiments on Charades-CON with TimeChat.} \textbf{(A)}: Performance evaluation across different ratios of event and temporal verification queries. The total queries are fixed at 12k, matching the \# of grounding queries in the original dataset.
    \textbf{(B)}: Performance evaluation across varying verification query scales. 100\% scale corresponds to the performance of VTune reported in the paper. The number of unique words is a proxy for dataset diversity.
    }
    \label{appendix: VTune}
\end{figure}

\noindent\textbf{How does VTune improve grounding?} We speculate that VTune enhances grounding in multiple ways. Fig~\ref{appendix: VTune}-(A) shows that verification tuning alone improves grounding performance by approximately 20\% (from an initial 30.5), suggesting that verification tasks inherently benefit grounding. This is likely because verification tasks are more general and align with the prior QA capabilities of LLMs, whereas fine-tuning for grounding may override this knowledge due to requiring task-specific timestamp templates, resulting in poor consistency. Additionally, Fig~\ref{appendix: VTune}-(B) highlights the importance of dataset scale and diversity, indicating that verification queries serve as effective data augmentation, expanding linguistic coverage. However, excessive training may diminish performance.

\begin{figure*}[t]
        \centering
        \includegraphics[width=0.99\linewidth]{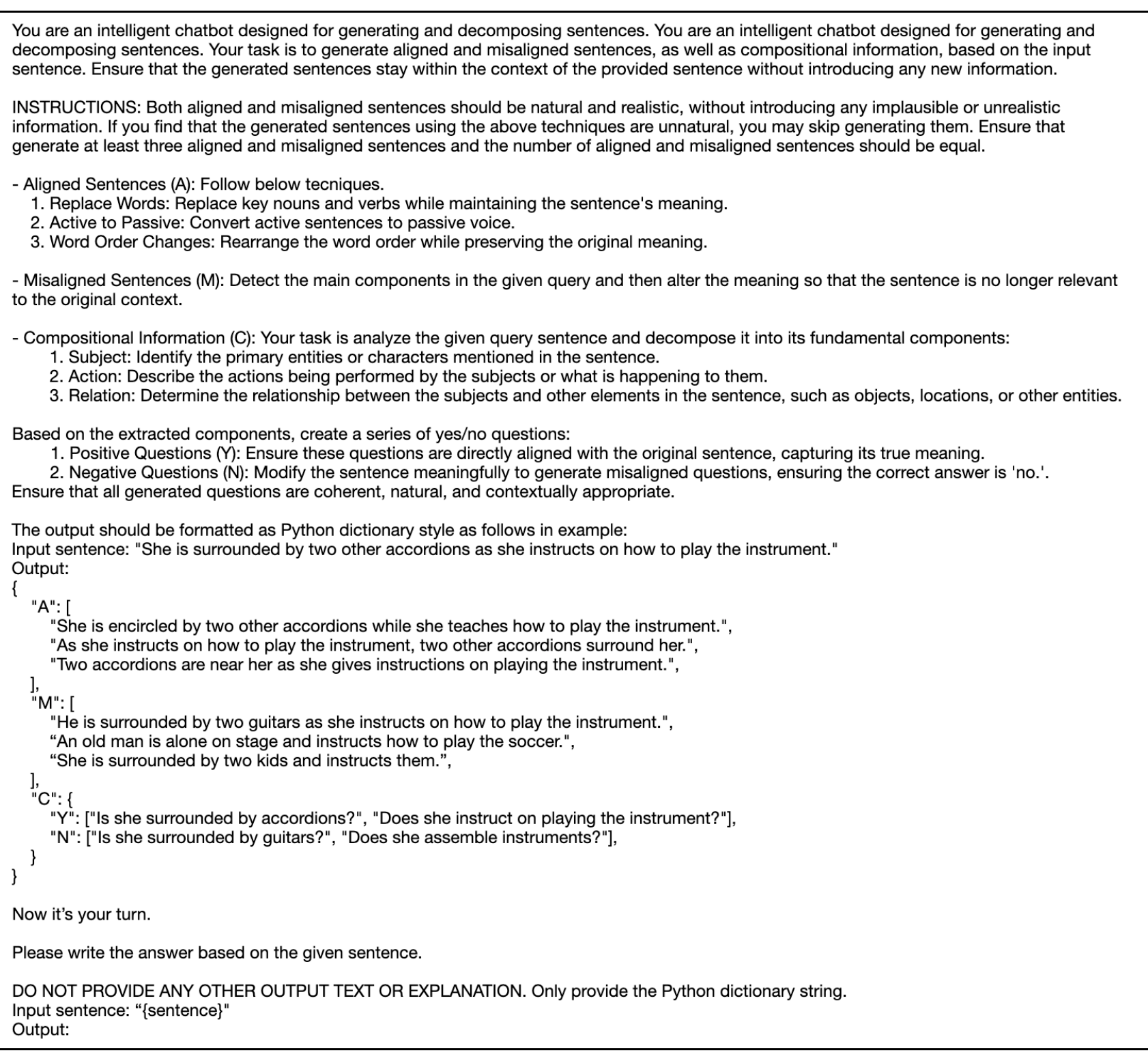}
    \caption{
    The prompt for generating aligned, misaligned, and compositional queries.
    }
    \label{fig: prompt to construct evaluation sets}
\end{figure*}

\begin{figure*}[t]
        \centering
        \includegraphics[width=0.99\linewidth]{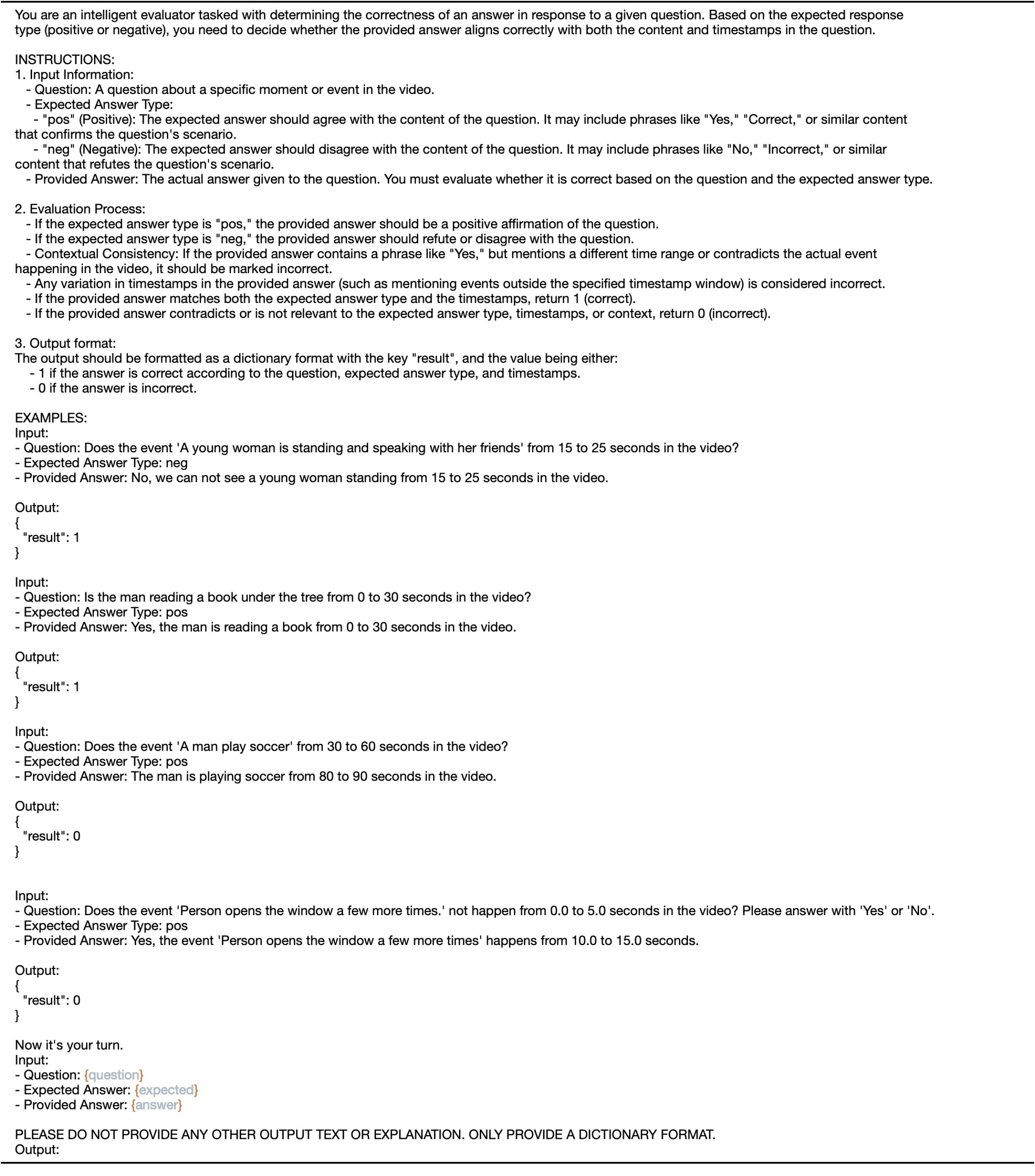}
    \caption{
    The prompt to assess the correctness of the model's answer for verification probes. Specifically, we ask GPT-4o-mini to assess whether the answer aligns with the expected answer.}
    \label{appendix_fig: prompt for evaluation}
\end{figure*}

\section{Prompts for GPT}
\label{appendix: prompt designs}
In this section, we provide detailed prompts that are utilized in our study.
\begin{enumerate}
    \item We use the prompt in Figure~\ref{fig: prompt to construct evaluation sets} to generate aligned, misaligned, and compositional queries.

    \item We use the prompt in Figure~\ref{appendix_fig: prompt for evaluation} to evaluate the model's answer for our probes.
\end{enumerate}
\input{appendix_tables/qa_examples}

%% file: appendix_tables/dataset_details.tex
\begin{table}[t]
    \begin{center}
    \resizebox{\linewidth}{!}{
      \begin{tabular}{lccccc}
      \toprule
      Dataset & Duration & Moment & GT Query & Aligned Query &  Misaligned Query \\
      \midrule
        Charades-CON & 30.7s & 8.1s & 7.5 & 10.6 & 8.3 \\
        ActivityNet-CON & 132.4s & 39.6s & 12.2 & 12.4 & 11.3 \\
      \bottomrule
      \end{tabular}
      }
    \end{center}
    \caption{
    Statistics of each dataset. We provide the average number of each column. The numbers for the queries represent the average word count.
    }
    \label{tbl: dataset_details}
\end{table}

%% file: appendix_tables/templates_for_holistic_verification.tex
\begin{table}[t]
    \begin{center}
        \resizebox{\linewidth}{!}{
      \begin{tabular}{ll}
      \toprule
      \bf{Expected Answer} & \bf Templates \\ 
      \midrule
      \multirow{4}{*}{Yes} & Is the event $q$ present from $m$ in the video? \\
        & Is the event $q$ occurring from $m$ in the video? \\
        & Does the event $q$ happen from $m$ in the video? \\
        & Is the event $q$ included from $m$ in the video? \\ \midrule
      \multirow{4}{*}{No} & Is the event $q$ absent from $m$ in the video? \\
        & Is the event $q$ not present from $m$ in the video? \\
        & Does the event $q$ not happen from $m$ in the video? \\
        & Is the event $q$ missing from $m$ in the video? \\ \bottomrule
      \end{tabular}
      }
    \end{center}
    \caption{
    Templates for holistic verification. Given the query sentence $q$ and the model's predicted timestamp $m$, the above question templates can be used for holistic verification.
    }
    \label{tbl: templates for holistic verification}
\end{table}

%% file: appendix_tables/grounding_templates.tex
\begin{table}[t]
    \begin{center}
    \resizebox{\linewidth}{!}{
      \begin{tabular}{ll}
      \toprule
      \bf{Method} & \bf Template \\ 
      \midrule
      \text{[1]} & When does the event $q$ happen in the video? Please\\ 
      & only return its start time and end time. \\ \midrule

      \text{[2]} & Please find the visual contents in the video described \\ 
      & by a given event, determining its starting. The format \\ 
      & should be: 'start time - end seconds'. Now I will give\\ 
      & you the event: $q$. Please only return its start time and\\
      & end time.\\ \midrule
    
      \text{[3]} & Please answer when the event $q$ occurs in the video. \\
      & The output format should be: 'start - end seconds'.\\ 
      & Please return its start time and end time." \\ 
      
      \midrule
      TimeChat & Please find the visual event described by a sentence \\ 
      & in the video, determining its starting and ending times. \\
      & The format should be: 'The event happens in the 'start \\
      & time - end time'. For example, The event 'person turn \\
      & a light on' happens in the 24.3 - 30.4 seconds. Now I \\
      & will give you the textual sentence: $q$. Please return\\
      & its start time and end time. \\ \midrule
      
      VTimeLLM & During which frames can we see the event $q$?\\ \midrule
      
      VTG-LLM & Find the video segment that corresponds to the given\\ 
      & textual query $q$ and determine its start and end seconds. \\ \midrule

      GPT-4o & In which time frames does the event $q$ happen? Iden-\\ 
      & tify the frames where the event takes place and provide\\ 
      & the answer like 'start - end seconds.'. For example, if\\
      & you  think the given event occurs from 5 to 10 seconds\\ 
      & in the video, please return the 'The given happens in 5\\
      & - 10 seconds.' Please only include the timestamp in the\\ 
      & answer without any other content. \\ \midrule
      
      Gemini 1.5 Flash & When does the event $q$ happen in the video? Please \\
      & return its start and end times. \\
      \bottomrule
      \end{tabular}
      }
    \end{center}
    \caption{Prompt templates for temporal grounding. We designed the templates of [1], [2], and [3] for \vllm~, as well as specific templates for GPT-4o and Gemini. For TimeChat, VTimeLLM, and VTG-LLM, we follow the official grounding prompts. We replace the $q$ in the prompts with the query sentence.
    }
    \label{tbl: grounding prompt templates}
\end{table}

%% file: appendix_tables/grounding_results.tex
\begin{table}[t]
    \resizebox{\linewidth}{!}{
      \begin{tabular}{llccccc}
      \toprule
      \multirow{2}{*}{\bf Method} & \multirow{2}{*}{\bf LLM} & \multirow{2}{*}{\bf Prompt} & \multicolumn{2}{c}{\bf Charades-CON} & \multicolumn{2}{c}{\bf ActivityNet-CON} \\
      \cmidrule(lr){4-5} \cmidrule(lr){6-7}
      & & & R@1, 0.5 & R@1, 0.7 & R@1, 0.5 & R@1, 0.7 \\ \midrule
        \multirow{6}{*}{Video-LLaMA} & \multirow{3}{*}{Llama2-7B} 
        & [1] & \bf 15.8 & \bf 5.7 & 9.6 & 2.0 \\
        & & [2] & 0.2 & 0.1 & 0.1 & 0.1 \\
        & & [3] & 0.3 & 0.2 & 0.4 & 0.0 \\ 

       & \multirow{3}{*}{Vicuna-7B} 
       & [1] & \underline{14.2} & \underline{4.1} & \underline{12.8} & \underline{3.1} \\
       & & [2] & 8.4 & 1.5 & \bf 16.8 & \bf 5.3 \\
       & & [3] & 3.4 & 0.3 & 2.8 & 1.0 \\

       & Vicuna-13B & [2] & 8.0 & 2.6 & 14.7 & 8.3 \\
       \midrule
       
        \multirow{3}{*}{Video-LLaVA} & \multirow{3}{*}{Vicuna-7B}
        & [1] & \multicolumn{4}{c}{\textit{------------ Invalid output ------------}} \\
        & & [2] & \bf 9.6 & \bf 3.9 & \underline{13.3} & \underline{5.4} \\
        & & [3] & \underline{9.4} & \underline{3.2} & \bf 13.4 & \bf 6.5 \\
        \midrule
   
       \multirow{3}{*}{Video-ChatGPT} & \multirow{3}{*}{Vicuna-7B} 
       & [1] & 12.3 & 6.8 & 3.0 & 0.3 \\
        & & [2] & \underline{13.5} & \underline{7.3} & \bf 3.4 & \underline{0.7} \\
        & & [3] & \bf 14.4 & \bf 7.6 & \underline{3.3} & \bf 0.8 \\ \midrule

        \multirow{3}{*}{VideoChat2} & \multirow{3}{*}{Vicuna-7B}
        & [1]  & 7.2 & 2.1 & 10.5 & 4.9 \\
        & & [2]  & \multicolumn{4}{c}{\textit{------------ Invalid output ------------}} \\
        & & [3]  & \multicolumn{4}{c}{\textit{------------ Invalid output ------------}} \\
        \midrule

        \multirow{3}{*}{Video-LLaMA2} & \multirow{3}{*}{Mistral-7B}
        & [1]  & \multicolumn{4}{c}{\textit{------------ Invalid output ------------}} \\
        & & [2]  & \underline{17.6} & \underline{8.3} & \bf 10.4 & \bf 4.8 \\
        & & [3]  & \bf 20.0 & \bf 9.3 & \bf 10.4 & \underline{4.7} \\ \midrule
        
        VTG-LLM & Llama2-7B & - & 26.0 & 11.0 & 6.8 & 3.0 \\ 
        VTimeLLM & Vicuna-7B & - & 27.3 & 11.4 & 31.9 & 17.0 \\
        TimeChat & Llama2-7B & - & 30.5 & 13.5 & 4.6 & 2.0 \\ 
        GPT-4o & - & - & 28.5 & 12.3 & 26.8 & 12.0 \\ 
        Gemini 1.5 Flash & - & - & 34.6 & 16.4 & 37.8 & 26.9 \\
      \bottomrule
      \end{tabular}
      }
    \caption{Grounding performance of \vllm~across various experiment settings. The number in the Prompts column denotes the prompts in Table~\ref{tbl: grounding prompt templates}. Note that the results are obtained from our evaluation sets, not the original test split. The \textit{Invalid output} represents when the model fails to include timestamps in its answer.
    }
    \label{tbl: finding the optimal setting for temporal grounding}
\end{table}

%% file: appendix_tables/grounding_GPT4o.tex
\begin{table}[t]
    \resizebox{\linewidth}{!}{
      \begin{tabular}{lccccc}
      \toprule
      \multirow{2}{*}{\bf Method} & \multirow{2}{*}{\bf \# Frames} & \multicolumn{2}{c}{\bf Charades-STA} & \multicolumn{2}{c}{\bf ActivityNet-Captions} \\
      \cmidrule(lr){3-4} \cmidrule(lr){5-6}
      & & R@1, 0.5 & R@1, 0.7 & R@1, 0.5 & R@1, 0.7 \\ \midrule
        \multirow{3}{*}{GPT-4o} & 10 & \bf 29.0 & \bf 16.0 & \bf 33.0 & \bf 13.0 \\
         & 32 & \underline{28.0} & \underline{10.0} & \underline{23.0} & \underline{9.0} \\
         & 64 & 13.0 & 9.0 & 18.0 & 5.0 \\
      \bottomrule
      \end{tabular}
      }
    \caption{Grounding performance of GPT-4o across different video frame inputs.
    For efficiency, we randomly sample 200 videos from Charades-STA and ActivityNet-Captions and find the acceptable grounding performance of GPT-4o.
    }
    \label{tbl: GPT-4o grounding}
\end{table}

%% file: appendix_tables/grounding_with_IoUs.tex
\begin{table*}[t]
    \resizebox{\linewidth}{!}{
      \begin{tabular}{lcccccccccccc}
      \toprule
      \multirow{2}{*}{\bf Method} & \multicolumn{6}{c}{\bf Charades-CON} & \multicolumn{6}{c}{\bf ActivityNet-CON} \\ 
       \cmidrule(lr){2-7} \cmidrule(lr){8-13} 
        & Ground & Ground (0.7) & R-Ground & R-Ground (0.7) & S-Ground & S-Ground (0.7) & Ground & Ground (0.7) & R-Ground & R-Ground (0.7) & S-Ground & S-Ground (0.7) \\ 
        \midrule
        \textcolor{gray}{\textit{Open-source (general-purpose)}} \\
         VideoChat2 & 7.2 & 2.1 & 5.4 (74.5) & 4.4 (60.7) & 1.0 (13.7) & 0.3 (3.9) & 10.5 & 4.9 & 8.7 (82.8) & 7.9 (75.6) & 0.6 (6.0) & 0.1 (1.3)  \\


        Video-LLaVA & 9.4 & 3.2 & 7.6 (80.8) & 7.4 (78.7) & 2.8 (30.3) & 0.4 (4.5) & \underline{13.4} & \underline{6.5} & \underline{10.0} (74.5) & \underline{9.4} (70.3) & 3.1 (23.0) & 1.0 (7.3) \\

        Video-LLaMA & 14.2 & 4.1 & 10.6 (74.9) & 9.3 (65.6) & 5.3 (37.6) & 8.3 (5.9) & 12.8 & 3.1 & 8.5 (66.8) & 8.1 (63.9) & \underline{7.2} (56.8) & \underline{1.6} (12.5) \\
        
        Video-ChatGPT & 14.4 & 7.6 & 12.8 (89.2) & 12.7 (87.9) & 1.3 (8.8) & 0.4 (2.9) & 3.3 & 0.8 & 2.8 (84.0) & 2.7 (82.6) & 0.1 (4.1) & 0.1 (4.1)  \\

        Video-LLaMA2 & 20.0 & 9.3 & 16.8 (83.8) & 16.6 (83.1) & 3.8 (19.0) & 1.1 (5.6) & 10.4 & 4.7 & 8.2 (78.6) & 7.4 (70.9) & 1.5 (14.8) & 0.3 (2.7)  \\
        \midrule
        \textcolor{gray}{\textit{Open-source (time-aware)}} \\
         VTG-LLM & 26.0 & 11.0 & 16.1 (62.1) & 11.4 (43.8) & \textbf{8.3} (32.0) & \textbf{2.5} (9.7) & 6.8 & 3.0 & 5.3 (78.0) & 3.1 (45.0) & 0.2 (3.0) & 0.1 (1.0)  \\

        VTimeLLM & \underline{27.3} & \underline{11.4} & \underline{22.7} (83.2) & 22.2 (81.3) & \underline{7.3} (26.9) & \underline{2.2} (8.2) & \bf 31.9 & \bf 17.0 & \textbf{26.4} (82.7) & \textbf{25.0} (78.5) & \underline{8.9} (27.9) & \textbf{3.6} (11.2)  \\
        
        TimeChat & \bf 30.5 & \bf 13.5 & \textbf{25.0} (82.1) & \underline{20.2} (66.2) & 5.6 (18.5) & 1.3 (4.1) & 4.6 & 2.0 & 2.9 (64.1) & 2.7 (58.5) & 1.0 (21.2) & 0.3 (7.5)  \\
         \midrule
        \textcolor{gray}{\textit{Closed-source}} \\
        \rowcolor{lightgray}
        
        GPT-4o & 28.5 & 12.3 & 21.2 (74.3) & 12.7 (44.6) & 9.3 (32.8) & 2.8 (9.9) & 26.8 & 12.0 & 18.1 (67.5) & 12.9 (48.3) & 10.4 (38.8) & 3.4 (12.5)  \\
        \rowcolor{lightgray}
        Gemini 1.5 Flash & 34.6 & 16.4 & 29.7 (85.7) & 24.8 (71.7) & 24.8 (71.7) & 13.5 (39.1) & 37.8 & 26.9 & 30.8 (81.4) & 27.6 (72.9) & 24.8 (65.6) & 14.8 (39.2)  \\
      \bottomrule
      \end{tabular}
      }
    \caption{Grounding and consistency results of models with different IoU thresholds. The number in brackets represents the IoU threshold. }
    \vspace{-2mm}
    \label{tbl: grounding and consistency across different IoUs}
\end{table*}

%% file: appendix_tables/hyperparameters.tex
\begin{table}[t]
    \begin{center}
    \resizebox{\linewidth}{!}{
      \begin{tabular}{lcccc}
      \toprule
      \multirow{2}{*}{\bf Hyper-parameter} & \multicolumn{2}{c}{\bf Instruction Tuning} & \multicolumn{2}{c}{\bf Event Temporal Verification Tuning} \\ 
      \cmidrule(lr){2-3} \cmidrule(lr){4-5}
      & Charades-STA & ActivitNet-Captions & Charades-STA & ActivitNet-Captions \\
      \midrule
        Epochs & 3 & 3 & 3 & 1 / 3 \\ 
        Batch size & 1 & 1 & 1 & 1 \\
        Learning rate & 3e-5 & 3e-5 & 3e-5 & 3e-5 \\
        Warm-up learning rate & 1e-6 & 1e-6 & 1e-6 & 1e-6 \\ \midrule
        Iters-per-epochs & 3102 & 24811 & 9332 & 51377 \\
        Warmup-steps & 3102 & 14916 & 9332 & 25688 \\
        
      \bottomrule
      \end{tabular}
      }
    \end{center}
    \caption{
    Hyper-parameters for each tuning method. While we train 3 epochs for Video-LLaMA with \method~on ActivityNet-Captions, we empirically find that TimeChat only needs 1 epoch for training. Except for this case, we use the same hyper-parameters in both models.
    }
    \label{tbl: hyperparameters}
\end{table}

%% file: appendix_tables/statistics_of_queries.tex
\begin{table}[t]
    \begin{center}
    \resizebox{\linewidth}{!}{
      \begin{tabular}{lcccc}
      \toprule
      \bf Dataset & \bf G & \bf E & \bf T & \bf Total\\
      \midrule
        Charades-STA & 12408 & 76145 & 10691 & 99244 \\ 
        ActivityNet-Captions & 37330 & 138340 & 29840 & 205510 \\ 
      \bottomrule
      \end{tabular}
      }
    \end{center}
    \caption{
    The number of grounding and verification queries in each dataset. The following columns represent a specific query type used: G for Grounding, E for Event Verification, and T for Temporal Verification.
    }
    \label{tbl: the number of queries}
\end{table}

%% file: appendix_tables/consistency_across_different_prompts.tex
\begin{table*}[t]
    \resizebox{\linewidth}{!}{
      \begin{tabular}{llcccccccccccc}
      \toprule
      \multirow{2}{*}{\bf Method} & \multirow{2}{*}{\bf LLM} & \multirow{2}{*}{\bf Prompt} & \multirow{2}{*}{\bf \# Frames} & \multicolumn{5}{c}{\bf Charades-CON} & \multicolumn{5}{c}{\bf ActivityNet-CON} \\ 
       \cmidrule(lr){5-9} \cmidrule(lr){10-14} 
        & & & & Ground & R-Ground & S-Ground & H-Verify & C-Verify
        & Ground & R-Ground & S-Ground & H-Verify & C-Verify \\ 
        \midrule
        \multirow{3}{*}{Video-LLaMA} & Llama2-7B & [1] & 8 
        & \bf 15.8 & \textbf{12.8} (81.2) & \underline{5.2} (33.2) & \underline{7.2} (46.2) & \textbf{7.9} (50.3) & 9.6 &8.2 (85.7) & 4.6 (48.7) & 4.8 (50.8) & 4.8 (50.7) \\

        & Vicuna-7B & [1] & 8 &
        \underline{14.2}& \underline{10.6} (74.9) & \textbf{5.3} (37.6) & \textbf{7.5} (53.3) & \underline{7.3} (51.7) &12.8&8.5 (66.8) & \underline{7.2} (56.8) & \underline{7.3} (57.5) & \underline{7.5} (58.9) \\
        
        & Vicuna-7B & [2] & 8 
        & 8.4 & 5.9 (70.0) & 2.9 (35.0) & 4.5 (54.0) & 4.6 (55.0) & \bf 16.8 & \textbf{14.5} (86.6) & \textbf{9.1} (54.1) & \textbf{9.8} (58.2) & \textbf{9.9} (59.1) \\
        
        & Vicuna-13B & [2] & 8 
        & 8.0 & 6.8 (85.9) &0.2 (3.5) &3.8 (48.3) &3.9 (49.5) &\underline{14.7} & \underline{12.8} (87.4) &1.0 (7.1) & 6.5 (44.6) & 7.1 (48.4) \\ \midrule
        
        \multirow{2}{*}{Video-LLaVA} & Vicuna-7B & [2] & 8 
        & \bf 9.6& \textbf{8.0} (84.3) & 2.4 (25.0) & 4.8 (50.8) & \textbf{4.8} (50.3) 
        &13.3& \textbf{10.4} (78.6) & \textbf{3.8} (28.9) & \textbf{6.4} (48.4) &6.8 (51.5) \\
        & Vicuna-7B & [3] & 8
        & 9.4 & 7.6 (80.8) & \textbf{2.8} (30.3) & \textbf{5.0} (52.8) & 4.7 (50.0) 
        &\textbf{13.4} & 10.0 (74.5) & 3.1 (23.0) & 6.3 (46.9) & \textbf{7.0} (52.3) \\ \midrule

        \multirow{2}{*}{Video-ChatGPT} & Vicuna-7B & [2] & 100
        & 13.8&12.0 (87.5) & 1.1 (8.3) & \textbf{6.6} (48.5) &6.9 (50.2) 
        &\textbf{3.4} & 2.7 (80.9) & \textbf{0.1} (4.0) & \textbf{1.7} (51.9) & \textbf{1.7} (51.6) \\
        & Vicuna-7B & [3] & 100
        & \bf 14.4 & \textbf{12.8} (89.2) & \textbf{1.3} (8.8) & 6.5 (44.8) & \textbf{7.2} (50.0) 
        & 3.3 & \textbf{2.8} (84.0) & \textbf{0.1} (4.1) & \textbf{1.7} (50.4) & 1.6 (49.2) \\ 
        \midrule

        \multirow{2}{*}{Video-LLaMA2} & Mistral-7B & [2] & 8 
        & 17.6&14.7 (83.7) &3.5 (20) &9.1 (52.1) &9.2 (52.8) 
        & \bf 10.4&7.8 (75.6) &1.4 (13.5) &5.3 (51.5) &5.6 (54.1) \\
        & Mistral-7B & [3] & 8 
        & \textbf{20.0} & \textbf{16.8} (83.8) & \textbf{3.8} (19.0) & \textbf{10.3} (51.5) & \textbf{10.6} (52.9) & \textbf{10.4} & \textbf{8.2} (78.6) & \textbf{1.5} (14.8) & \textbf{5.4} (52.4) & \textbf{5.7} (54.7) \\ 
        
        \bottomrule
        \end{tabular}
        }
        \caption{Consistency evaluation of \vllm~across different settings. While the model's performance varies across different prompts for temporal grounding and LLM backbones, there are no significant differences between them.}
        \label{tbl: consistency of vllm across different settings}
\end{table*}

%% file: appendix_tables/rephrased_grounding.tex
\begin{table}[t]
    \resizebox{\linewidth}{!}{
    \begin{tabular}{lccc>{\columncolor[gray]{0.9}}cccc>{\columncolor[gray]{0.9}}c}
      \toprule
      \multirow{2}{*}{\bf Method} & \multicolumn{4}{c}{\bf Charades-CON} & \multicolumn{4}{c}{\bf ActivityNet-CON} \\
      \cmidrule(lr){2-5} \cmidrule(lr){6-9}
      & WR & AP & WO & Avg & WR & AP & WO & Avg \\ \midrule
        VideoChat2 & \bf 76.2 & 75.3 & \underline{75.7} & 75.7 
        & \bf 86.0 & \underline{84.7} & 81.0 & 83.9 \\
        
        Video-LLaVA & \bf 83.7 & 74.7 & \underline{75.5} & 77.9 
        & \bf 76.4 & 70.0 & 69.7 & 72.0 \\

                Video-LLaMA & \underline{74.3} & \bf 75.0 & 65.1 & 71.4 
        & \bf 89.1 & \underline{85.3} & 85.2 & 86.5 \\
        
        Video-ChatGPT & \bf 92.8 & 86.6 & 90.8 & 90.0 
        & \underline{88.5} & 88.5 & \bf 88.8 & 88.6 \\
        
        Video-LLaMA2 & \bf 88.0 & 82.0 & \underline{82.8} & 84.2 
        & \bf 82.6 & \underline{71.3} & 69.9 & 74.6 \\
        
        VTG-LLM & \bf 56.8 & \underline{53.8} & 53.8 & 54.8 
        & \bf 77.2 & 60.4 & 61.3 & 66.3 \\
        
        VTimeLLM & \bf 77.3 & \underline{74.9} & 73.4 & 75.2 
        & \bf 63.3 & 62.8 & 63.3 & 63.1 \\
        
        TimeChat & \underline{83.8} & \bf 84.7 & 80.4 & 82.9 
        & \bf 84.4 & \underline{80.5} & 78.6 & 81.1 \\
        
        GPT-4o & 60.0 & \bf 64.7 & \underline{61.9} & 62.2 & 56.9 
        & \bf 58.5 & \underline{57.3} & 57.5 \\
        
        Gemini 1.5 Flash & \underline{75.4} & \bf 77.4 & 74.9 
        & 75.9 & \bf 76.0 & \underline{71.4} & 71.4 & 72.9 \\
      \bottomrule
      \end{tabular}
      }
    \caption{Rephrased grounding results across three modification rules. WR: Word Replacement, AP: Active-to-Passive conversion, and OC: Word Order change. We report mIoU for rephrased grounding. Most models generally achieve a high score for WR.
    }
    \label{tbl: mIoU of rephrased grounding}
\end{table}

%% file: appendix_tables/qa_examples.tex
\begin{table*}[t]
    \begin{center}
    \resizebox{\linewidth}{!}{
      \begin{tabular}{ll}
      \toprule
      \bf{Method} & \bf Question-Answer pairs \\ 
      \midrule
      Video-LLaMA 
      & Q: When does the event `person puts bag of groceries on table.'? Please only return its start time and end time. 
      \\
      & A: The event 'person puts bag of groceries on table' happens at 0.0 seconds.  It lasts for 19.2 seconds.
      \\ 
      & Q: Is the person running into the room from 0.0 to 19.2 seconds in the video? 
      \\
      & A: Yes.
      \\ \midrule
      
      Video-LLaVA 
      & Q: Please answer when the event `He keeps adding paint along the way, sometimes directly to the canvas.' occurs in the video.
      \\
      &~~~~~The output format should be: `start - end seconds'. Please return its start time and end time. 
      \\ 
      & A: 0:45 - 0:65 
      \\
      & Q:  Does the event `man is taking off a violin from the case and gives it to a man.' not happen from 0.0 to 36.0 seconds in the video? 
      \\
      & A: Yes, the event 'man is taking off a violin from the case and gives it to a man' does not happen from 0.0 to 36.0 seconds in the video. 
      \\ \midrule
      
      Video-ChatGPT 
      & Q: Please answer when the event 'The man then instructs the man holding the rope to pull the row down a few times and he's talking
      \\
      &~~~~~the whole time.' occurs in the video. The output format should be: 'start - end seconds'. Please return its start time and end time. 
      \\
      & A: The event 'The man then instructs the man holding the rope to pull the row down a few times and he's talking the whole time.'
      \\
      &~~~~~occurs at 0:15 - 0:30. 
      \\
      & Q: Is the event 'The man subsequently instructs the individual holding the rope to draw the row down multiple times while he's speaking
      \\ 
      &~~~~~continuously.' present from 15 to 30 seconds in the video? 
      \\
      & A: Yes, the event is present from 15 to 30 seconds in the video. 
      \\ \midrule
      
      Video-Chat2 
      & Q: When does the event 'person closes the door a few more times.' occurs in the video? 
      \\ 
      &~~~~~Please return its start time and end time in seconds. The output format should be: 'start - end seconds'. 
      \\
      & A: The event 'person closes the door a few more times' occurs in the video at the start time of 00:00 and ends at 00:02. 
      \\
      & Q: Is the event 'The individual shuts the door several additional times.' occurring from 0 to 2 seconds in the video? 
      \\
      & A: The event of the individual shutting the door several times is happening from 0 to 2 seconds in the video. 
      \\ \midrule
      
      Video-LLaMA2 
      & Q: Please answer when the event 'A person is seen knitting close up and leads into a woman speaking.' occurs in the video. \\ 
      &~~~~~The output format should be: 'start - end seconds'. Please return its start time and end time.\\
      & A: The event 'A person is seen knitting close up and leads into a woman speaking.' takes place from 0 seconds to 11.9 seconds. \\
      & Q: Is the event 'Close up, a person is observed knitting, leading into a woman who is speaking.' not present from 0 to 11.9 seconds \\ 
      &~~~~~in the video? \\
      & A: Yes, the event 'Close up, a person is observed knitting, leading into a woman who is speaking.' is not present from 0 to 11.9 \\ 
      &~~~~~seconds in the video. \\ \midrule
      
      TimeChat 
      & Q: Please find the visual event described by a sentence in the video, determining its starting and ending times. The format should be:\\ 
      &~~~~~'The event happens in the start time - end time'. For example, The event 'person turn a light on' happens in the 24.3 - 30.4 seconds.\\ 
      &~~~~~Now I will give you the textual sentence: 'person starts to run out of the room.' Please return its start time and end time. \\
      & A: The given query happens in 20 - 25 seconds. \\
      & Q: Is the event 'Out of the room, the person starts to run.' occurring from 20.0 to 25.0 seconds in the video? Please answer with 'Yes'\\
      &~~~~~or 'No'. \\
      & A: Yes. \\ \midrule
      
      VTimeLLM 
      & Q: During which frames can we see the event 'The man then instructs the man holding the rope to pull the row down \\ 
      &~~~~~a few times and he's talking the whole time.'?\\
      & A: The event 'The man then instructs the man holding the rope to pull the row down a few times and he's talking \\ 
      &~~~~~the whole time.' can be seen from 17 to 71. \\
      & Q: Is the event 'The woman then instructs the man holding the rope to pull the row up a few times and she's \\ 
      &~~~~~silent the whole time.' occurring from 17 to 71 frames in the video? \\
      & A: Yes, that's correct. \\ \midrule

      VTG-LLM 
      & Q: Find the video segment that corresponds to the given textual query 'person puts a bag on the sofa.' and determine \\ 
      &~~~~~~its start and end seconds. \\
      & A: 0020.0 - 0025.0 seconds, person puts a bag on the sofa. \\
      & Q: Is the event 'person puts a bag on the sofa' occurring from 20.0 to 25 seconds in the video? Please answer with 'Yes' or 'No'. \\
      & A: 0020.0 -  0025.0  seconds, person puts a bag on the sofa.\\ 
      \bottomrule

      GPT-4o
      & Q: The video consists of 10 frames, with a total duration of 153.35 seconds. Each frame is sampled every 15.0 seconds. In which time \\
      &~~~~~~frames does the event He mixes some pain and paints part of a canvas, then does another color. happen? Identify the frames\\
      &~~~~~~where the event takes place and provide the answer like start - end seconds.. For example, if you think the given event occurs from \\
      &~~~~~~5 to 10 seconds in the video, please return the 'The given happens in 5 - 10 seconds.' Please only include the timestamp in the \\
      &~~~~~~answer without any other content. \\ 
      & A: The given event happens in 0 - 60 seconds. \\ \bottomrule

        Gemini 1.5 Flash
      & Q: When does the event 'a person walks through a doorway.' happen in the video? Please return its start and end times.\\
      & A: The person starts walking through the doorway at 00:07 and finishes at 00:24. \\
      & Q: Is the event 'a person walks through a doorway.' included from 7 to 24 seconds in the video? Please answer with Yes or No.\\
      & A: Yes.\\       
      \bottomrule
      \end{tabular}
      }
    \end{center}
    \vspace{-2.5mm}
    \caption{
    Example question-answer pairs of the models during our evaluation.
    }
    \label{tbl: qa examples}
\end{table*}